\documentclass[10pt,twocolumn,letterpaper]{article}

\usepackage{cvpr}              

\usepackage{paralist}
\usepackage{amsmath}
\usepackage{bbm}
\usepackage{multirow}
\usepackage{soul}
\usepackage[linesnumbered,ruled,vlined]{algorithm2e}
\usepackage{mathabx}
\usepackage{pifont}
\usepackage{stfloats}

\newcommand{\bsb}[1]{\boldsymbol{#1}}

\newcommand{\mbf}[1]{\mathbf{#1}}
\newcommand{\mac}[1]{\mathcal{#1}}
\newcommand{\diag}{\mathrm{diag}}

\newcommand{\udl}[1]{\underline{#1}}
\newcommand{\vspacex}{\vspace{-3.5mm}}

\newcommand{\abc}[1]{#1}

\newcommand{\NOTE}[1]{}
\newcommand{\lw}[1]{#1}
\newcommand{\lws}[2]{#2}
\newcommand{\lwn}[1]{#1}

\newcommand{\ANS}[1]{}
\newcommand{\CUT}[1]{}

\definecolor{cvprblue}{rgb}{0.21,0.49,0.74}
\usepackage[pagebackref,breaklinks,colorlinks,allcolors=cvprblue]{hyperref}

\def\paperID{14684} 
\def\confName{CVPR}
\def\confYear{2025}

\title{Point-to-Region Loss for Semi-Supervised Point-Based Crowd Counting}

\author{Wei Lin, \ Chenyang Zhao, \ and \ Antoni B. Chan\\
 Department of Computer Science, City University of Hong Kong\\
{\tt\small elonlin24@gmail.com, chenyzhao9-c@my.cityu.edu.hk, abchan@cityu.edu.hk}
}

\begin{document}
\maketitle

\begin{abstract}
Point detection has been developed to locate pedestrians in crowded scenes by training a counter through a point-to-point (P2P) supervision scheme. Despite its excellent localization and counting performance, training a point-based counter still faces challenges concerning annotation labor: hundreds to thousands of points are required to annotate a single sample capturing a dense crowd. In this paper, we integrate point-based methods into a semi-supervised counting framework based on pseudo-labeling, enabling the training of a counter with only a few annotated samples supplemented by a large volume of pseudo-labeled data. However, during implementation, the training encounters issues as the confidence for pseudo-labels fails to be propagated to background pixels via the P2P. To tackle this challenge, we devise a point-specific activation map (PSAM) to visually interpret the phenomena occurring during the ill-posed training. Observations from the PSAM suggest that the feature map is excessively activated by the loss for unlabeled data, causing the decoder to misinterpret these over-activations as pedestrians. To mitigate this issue, we propose a point-to-region (P2R) scheme to substitute P2P, which segments out local regions rather than detects a point corresponding to a pedestrian for supervision. Consequently, pixels in the local region can share the same confidence with the corresponding pseudo points. Experimental results in both semi-supervised counting and unsupervised domain adaptation highlight the advantages of our method, illustrating P2R can resolve issues identified in PSAM. The code is
available at \url{https://github.com/Elin24/P2RLoss}.
\vspace{-5mm}
\end{abstract}

\section{Introduction}
\label{sec:intro}

\begin{figure}[t]
	\centering
	\includegraphics[width=0.46\textwidth]{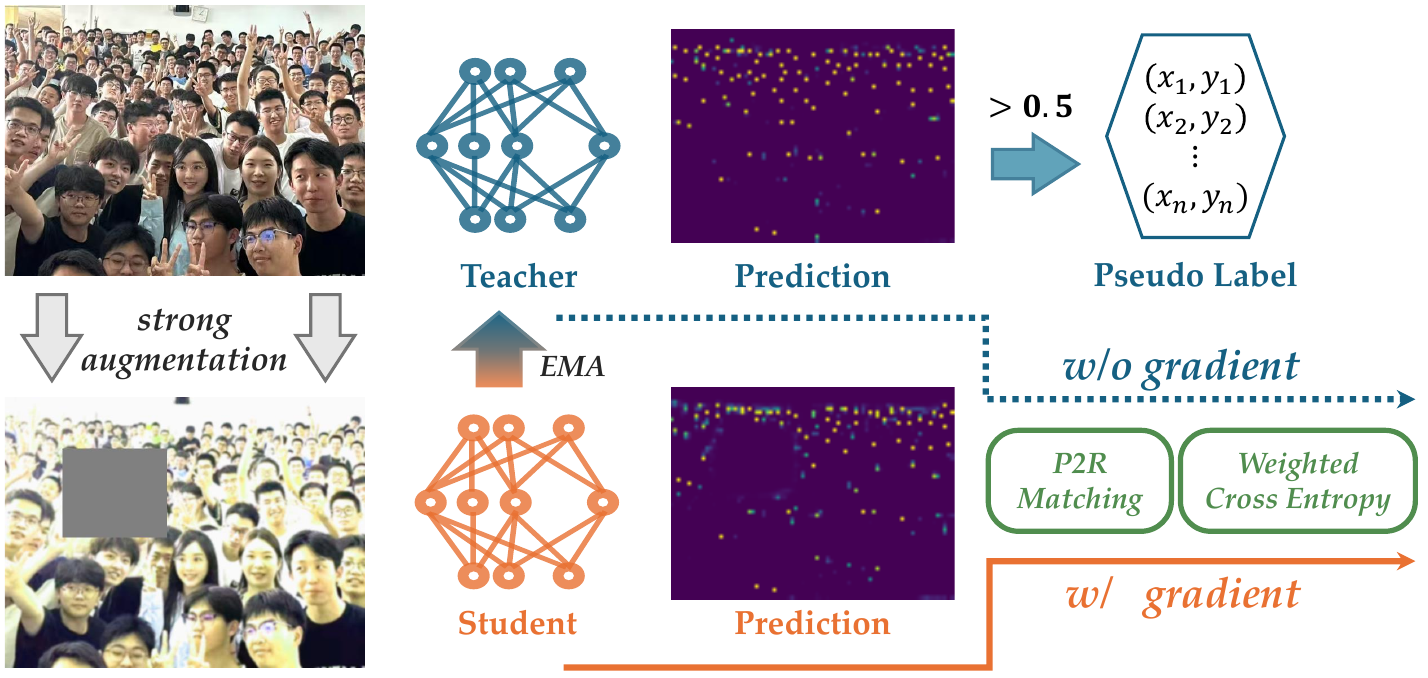}
	\caption{The workflow of semi-supervised point-based counting methods. The teacher model generates pseudo labels by extracting the foreground pixels, while the student model takes the corresponding strongly augmented image as input to construct the computation graph. The training loss between the pseudo label and the student's prediction involves two steps: the proposed P2R matching and the weighted cross-entropy computation.}
	\label{fig:fixmatch}
	\vspace{-5mm}
\end{figure}

Crowd scene analysis has been investigated for many years due to its significant applications in smart cities and urban safety~\citep{ldaec, groupdt, tbc, 8444178, alhothali2023anomalous, chen2017anchor, rccssam}. Crowd counting and localization in extremely dense crowd scenes, as the fundamental task in this field, have attracted considerable attention from researchers in computer vision~\citep{nwpu, gl, cline, bprc, cpbr, dcl}. 
Although most supervised crowd counting methods follow the scheme of density regression~\citep{mcnn, bl, gl, man, steerer}, there are also relevant studies implementing point detection for crowd localization and counting~\citep{p2p, cltr, pet}, due to its straightforwardness in indicating pedestrians' positions. These approaches use the Hungarian algorithm to perform point-to-point (P2P) matching between predicted point maps and ground-truth (GT) pedestrian annotations during training. Therefore, these matched pixels are considered as pedestrians' positions, while the others are designated as background for loss computation. This scheme successfully trains a point detector for crowd analysis, yielding both good localization and counting performance.

Great progress has been made in supervised counting, but the labor-intensive nature of annotating data has hindered the development of crowd understanding for a long time~\cite{otm, nwpu}. 
Addressing this issue, semi-supervised counting is gradually being developed, aiming to train a model using large amounts of unlabeled data alongside only a few annotated samples. 
Similar to supervised counting, most semi-supervised methods also follow the procedure of density regression. Notably, the method proposed in OT-M~\citep{otm} stands out by employing a simple self-training scheme based on mean-teacher~\citep{mt} and FixMatch~\citep{fixmatch}. However, there is no study exploring semi-supervised counting and localization using a point-detection method. To fill this gap, this paper explores an approach to train a point detector with limited annotated data.

As shown in Fig.~\ref{fig:fixmatch}, typical semi-supervised counting methods rely on pseudo-labeling, which generates pseudo labels to serve as learning targets for unlabeled samples using a model pre-trained on a small amount of annotated data. During training, there is a confidence score to determine whether a concerned prediction should be retained as a pseudo label.
However, if the P2P matching strategy is applied to compute loss within the pipeline presented in Fig.~\ref{fig:fixmatch}, there is no reasonable way to define the confidence for \emph{unmatched background} pixels, as P2P matching only cares about \emph{matched foreground} pixels. Specifically, a foreground pixel can be retained for training if its matched pseudo-label's foreground probability exceeds a predefined threshold, similar to the approach used in semi-supervised classification tasks~\citep{psdlab, fixmatch}. However, background pixels lack confidence scores since no pseudo-labels are assigned to them after matching. Experimental results also demonstrate that ignoring the background leads to the breakdown of semi-supervised training.

To address the aforementioned issue, we propose a gradient-based visual explanation method (akin to~\cite{cam, gcam, odamc, odamj}), termed point-specific activation map (PSAM) , to visualize regions of interest and the activation levels of detected points. Using PSAM, we observe that the activation values and coverage areas of foreground pixels continuously grow during semi-supervised training, causing the model to misclassify neighboring pixels as foreground.

Inspecting PSAMs reveals that each point in GT is implicitly associated with a local region rather than a single pixel in the prediction. Accordingly, we propose to replace the P2P with a point-to-region (P2R) strategy, which matches local regions to points in the pseudo-label. This allows all pixels within a matched region, including background pixels, to share the confidence of the corresponding pseudo-point. Experimental results and visualizations demonstrate that P2R can effectively train a semi-supervised point-based counter and achieve expected performance. Experimental results show that P2R effectively trains semi-supervised point-based counters with comparable performance to P2P but at significantly reduced computational cost. Unlike P2P~\cite{p2p}, P2R eliminates the need for the Hungarian algorithm by assigning each ground-truth or pseudo-point a local "dominated zone" and selecting a representative foreground pixel based on a predefined cost matrix.

The contributions of our paper are as follows:
\begin{compactitem}
	\item We formulate a pipeline of semi-supervised counting with a point-detection method, but find a problem leading the training breakdown.
	\item We propose Point-Specific Activation Map (PSAM) to visualize the regions of interest and the activation values of each point detected by the point-based crowd counter. Through PSAM, we observe that the surrounding pixels of each expected foreground pixel are over-activated, causing the decoder to recognizes these pixels as instances different from the concerned one.
	\item Based on insights from PSAM, we propose replacing the P2P with a P2R matching strategy for semi-supervised learning in point-based crowd counting. This not only makes the training process work as expected, but also brings additional benefits of reducing computational requirements.
	\item The proposed P2R achieves outstanding performance in several experiments, including semi-supervised counting and unsupervised domain adaptation. Ablation study shows that P2R excellently tackles the problem caused by P2P in semi-supervised counting.
\end{compactitem}

\section{Related Works}
\label{sec:reworks}
{\bf Object counting in dense scenes.} The development of deep learning has significantly influenced crowd counting. Prior to this, detecting body parts~\citep{tra1, tra2} was the common pipeline for crowd understanding in computer vision, although some methods implemented counting models without explicit object segmentation or tracking information~\citep{ppcm, bprc, cpbr}. Beyond specific crowd scenarios, class-agnostic counting~\citep{ltce, zsoc, spdcn, fpa, clipc, tfpoc} has also been developed to estimate the distribution of objects in the same category. Most recent counting methods implement neural networks following the density \abc{map} regression scheme~\citep{mcnn, switchcnn, csrnet, sfcn, sfcnp, dacl, kbdm, ccfd, sicn}. In addition to continuous innovations in model architecture design~\cite{csrnet, switchcnn, man}, new loss functions have also been developed, including Bayesian Loss~\cite{bl}, loss functions based on balanced~\citep{dmc} and unbalanced optimal transport~\cite{uotc, dmmc, gl}, and characteristic function loss computed in the frequency domain~\citep{chfl, gcfl}. A study analyzes the relationship between current loss functions from the perspective of proximal mapping~\citep{pml}.
Some methods also perform crowd analysis via instance segmentation~\cite{topo, iim} and point detection~\cite{p2p, cltr, pet, psl}. The latter is the scheme adopted in our semi-supervised model.

{\bf Semi-supervised counting.} Semi-supervised counting has attracted broad attention recently~\citep{l2r, sl2r, gp, irast, sua, dac, otm, mtp, hpl, p3net, cu}. L2R~\citep{l2r, sl2r} introduces a ranking rule for unlabeled crowd images based on the principle that cropped images should have smaller counts than full images. GP~\citep{gp} iteratively updates pseudo-labels using Gaussian Processes. IRAST~\citep{irast} and SUA~\citep{sua} enhance models by incorporating a segmentation branch. DAC~\citep{dac} uses learnable density agents to capture varying crowd densities. OT-M~\citep{otm} transforms density maps into point maps as pseudo-labels, achieving strong performance with a density-to-point loss function~\citep{gl}. Our method also employs pseudo-labeling but uses a point detector for counting, avoiding the need for density map transformations as in~\citep{otm}. Additionally, uncertainty is directly measured by the scores of extracted points, without any additional learnable structures like those in~\cite{sua}.

{\bf Visual Explanation.} Visualizing the importance of extracted image features is a straightforward way to interpret a model. The mainstream of visual explanation focuses on methods producing activation maps~\citep{cam, gcam, gcamp, scam, acam, odamc, odamj, eclip}, which are the products of feature maps from some intermediate layer and 
weight maps indicating the corresponding feature's importance. For example, 
Grad-CAM~\citep{gcam, gcamp} defines the weights using the corresponding gradient maps, generated via backpropagation. Beyond the focus on classification tasks, ODAM~\citep{odamc, odamj} extends gradient-based CAM to object detection, visualizing the local impact of features on the detector’s decisions. 
In this paper, we propose PSAM to visualize how the activation map changes for each predicted point in an ill-posed semi-supervised counting framework. Based on the observations from PSAM, we design a P2R matching strategy to avoid the over-activation of pixels in local regions, which effectively resolves the drawbacks of P2P and enables the successful training of a semi-supervised point detector.

\section{Point Counter in Semi-Supervised Counting}

This section formulates a simplified point-based counter by removing the regression branch in P2PNet~\citep{p2p}, and then presents the training scheme and challenges when training with labeled and unlabeled data, respectively.

\subsection{Preliminary on A Simplified Point Counter}

\lwn{Common point-based crowd counters like P2PNet~\citep{p2p} contain a classification and a regression branch. However, we find a simplification can be achieved by \abc{removing} the offset regression branch. On one hand, the crowd count can be estimated by counting foreground pixels with scores greater than 0.5, a process that does not involve the regression branch. On the other hand, pixel coordinates alone are sufficient to locate pedestrians, as demonstrated in STEERER~\citep{steerer} and GL~\citep{gl}, which use the coordinates of local maxima to indicate pedestrians locations.}
	
\lwn{As a result, P2PNet can be simplified to:
\begin{align}
	\mac{F}(I, \Theta_f) &\rightarrow F \in \mathbb{R}^{c\times h \times w}, \label{eq:encoder} \\
	\mac{D}(F, \Theta_d) &\rightarrow D \in \mathbb{R}^{h \times w} \label{eq:p2pnet} \\ 
	&\rightarrow \mac{P} = \left\{\bsb{p} \in \mathbb{R}^n, \bsb{x} \in \mathbb{R}^{n \times 2}\right\}, \label{eq:p2pout}
\end{align}
where $\mac{F}$ is the image encoder with learnable parameters $\Theta_f$, and $\mac{D}$ denotes the density decoder with learnable parameter $\Theta_d$. Notably, the offset regression branch is excluded in this simplified formulation. The pixel set $\bsb{p} \in \mathbb{R}^n$ represents the flattened values in $D$, and its corresponding coordinate set is denoted as $\bsb{x} \in \mathbb{R}^{n \times 2}$, where $n = h \times w$.}

\subsection{Training with Labeled Data via P2P Matching}
\label{sec:tld}

\lwn{The GT with $m$ points can be denoted as $\bsb{x}' \in \mathbb{R}^{m \times 2}$, where ${\bsb{x}'}_{[j]} \in \mathbb{R}^2$ represents the location of the $j$-th pedestrian. Using $\mac{P}$ and $\bsb{x}'$, the loss computation involves two parts: P2P matching and binary cross entropy (BCE) calculation. }

\lwn{The former aims to obtain a matrix-based solution $\mbf{M} \in \{0, 1\}^{n\times m}$ that minimize the bipartite-graph matching cost between $\bsb{x}$ and $\bsb{x}'$ within the cost matrix $\mbf{C} \in \mathbb{R}^{n \times m}$:
\begin{equation}
	\mbf{C}_{[ij]} = \tau \|\bsb{x}_{[i]} - {\bsb{x}'}_{[j]}\|_2 - \mac{S}(\bsb{p}_{[i]}), \label{eq:cost}
\end{equation}
where $\mac{S}(\cdot)$ is defined as the inverse sigmoid function $\mac{S}(p) = -\log(1 / p - 1)$, rather than the identity operator used in the vanilla P2PNet~\cite{p2p}, for improved performance.}

To compute BCE loss, the learning objective of $\bsb{p}$ is formulated as $\hat{\bsb{p}} = \mbf{M}\mbf{1}_m$, and the BCE loss is calculated for optimization as follows:
\begin{align}
	\mac{L}_{l} = - \lambda \hat{\bsb{p}}^\top \log \bsb{p} - (\mbf{1}_n - \hat{\bsb{p}})^\top \log (\mbf{1}_n - \bsb{p}) \label{eq:bce}
\end{align}
where $\lambda$ is a re-weighting factor for matched pixels~\citep{p2p}.

\subsection{Problem in Training with Unlabeled Image}

Here we consider training with unlabeled data following  the \abc{standard} pseudo-labeling procedure~\cite{psdlab, fixmatch}. As illustrated in Fig.~\ref{fig:fixmatch}, it requires the cooperation of a student and teacher model that share the same structure described in \eqref{eq:encoder}$\sim$\eqref{eq:p2pout}. The difference lies in that the teacher model is updated via exponential moving average (EMA)~\citep{mt}, while the student is optimized via gradient descending. During training, the teacher and student process the weakly and strongly \lw{data-}augmented versions of an unlabeled image, producing $\mac{P}_t = \{\bsb{p}_t, \bsb{x}_t\}$ and $\mac{P}_s = \{\bsb{p}_s, \bsb{x}_s\}$, respectively. After that, foreground points are extracted from $\mac{P}_t$ to construct pseudo-labels:
\begin{align}
	\mac{P}'_t = \{\bsb{p}'_t, ~\bsb{x}'_t\} = \big\{{\bsb{p}_t}_{[j]}, ~{\bsb{x}_t}_{[j]} ~\big|~ {\bsb{p}_t}_{[j]} > 0.5 \big\}, \label{eq:sdlabel}
\end{align}
\lwn{where $\bsb{x}'_t$ is treated as the pseudo hard label to estimate the matching matrix $\mbf{M}_{st}$ via Hungarian algorithm}, and the loss is obtained by substituting  $\hat{\bsb{p}}_t = \mbf{M}_{st}\mbf{1}_m$ and $\bsb{p}_s$ into \eqref{eq:bce}.

However, some pseudo-labels may not be reliable. A common solution is to only consider points in $\mac{P}'_t$ with high scores~\cite{psdlab, fixmatch}, \textit{i.e.}, using a confidence vector $\bsb{\zeta}$ to mask out elements whose score is below a threshold $\eta$:
\begin{equation}
	\bsb{\zeta} = \mathbbm{1}\left(\bsb{p}'_t > \eta\right) ~\xrightarrow{\text{map to } \hat{\bsb{p}}_t \text{ by } \mbf{M}_{st}}~ \bsb{z} = \mbf{M}_{st}\bsb{\zeta}. \label{eq:conf}
\end{equation}
\lw{Note $\eta$ is greater than 0.5 and \eqref{eq:conf} only let pseudo points with score greater than $\eta$ be reliable. Subsequently, $\mbf{Z} = \diag(\bsb{z})$ is incorporated into the BCE loss for supervision:}
\begin{equation}
	\mac{L}_{u} = - \lambda \hat{\bsb{p}}_t^\top\mbf{Z}\log\bsb{p}_s - (\mbf{1}_n - \hat{\bsb{p}}_t)^\top\mbf{Z}\log(\mbf{1}_n - \bsb{p}_s). \label{eq:bce_z}
\end{equation}
However, the second item of \eqref{eq:bce_z} in P2P is identically zero:
\begin{equation}
	\mac{L}_{u\texttt{[P2P]}} = - \lambda \hat{\bsb{p}}_t^\top\mbf{Z}\log\bsb{p}_s - 0, \label{eq:bce_u}
\end{equation}
regardless of $\bsb{\zeta}$ (embedded in $\mbf{Z}$). Consequently, no gradients are back-propagated from the losses associated with background pixels, which is ill-posed. Utilizing \eqref{eq:bce_u} as the loss function is similar to training a two-class classifier with only samples from the category ``1'', which will lead the trained model to output ``1'' regardless of the input.

\NOTE{what is the problem then? expand on it more. }\ANS{I suppose that \eqref{eq:bce_u} is similar to train a classifier with samples in one category. Thus the model will just output the category involved in training regardless of the input.}

\section{Visual Explanation with PSAM}

In addition to the theoretical analysis, we implement a framework that supervises a model using labeled and unlabeled data via \eqref{eq:bce} and \eqref{eq:bce_u}, respectively, to empirically investigate the behavior of P2P-based semi-supervised counting. Subsequently, PSAM is designed to visualize how the prediction is influenced by \eqref{eq:bce_u} during training. Thus, we can design an effective confidence scheme accordingly to counteract the effects revealed by PSAM, thereby enabling semi-supervised learning to function as expected.

\subsection{PSAM: Point-Specific Activation Map}

An activation map is a heat map with the same spatial resolution as the image features, where local regions with relatively high values are interpreted as being discriminative \abc{(informative)} for the current output~\citep{cam, gcam, odamc}. Most techniques obtain activation maps by linearly aggregating features with their corresponding gradients, as 
\cite{gradweight, gradweight2} have shown that the partial derivative with respect to a specific element can reflect its contribution to the final prediction.

Unlike \cite{cam, gcam}, which compute gradients at the image level, crowd counting requires a point-specific activation map (PSAM) to visually explain the regions of interest for every element in $\bsb{p}$ of \eqref{eq:p2pout}. Thus, the Jacobian matrix of $\bsb{p}$ with respect to image feature $F$ is required:
\begin{align}
	\mbf{J}_\mac{D} = \tfrac{\partial \bsb{p}}{\partial F} = \big[\nabla\bsb{p}_{[1]}, \nabla\bsb{p}_{[2]}, \cdots, \nabla\bsb{p}_{[n]}\big]^\top. \label{eq:jcb}
\end{align}
Afterwards, the $q$-th heat map $H_{[q]}$ in PSAM is derived by filtering out negative influences on the aggregated feature:
\begin{align}
	H_{[q]} = \max\big(\sum\nolimits_{k=1}^c \nabla\bsb{p}_{[q]} \odot F, 0\big) \in \mathbb{R}^{h \times w}, \label{eq:hmap}
\end{align}
where the summation operator is applied along the channel dimension, and $\odot$ denotes the Hadamard product.

\begin{figure}[t]
	\centering
	\includegraphics[width=0.47\textwidth]{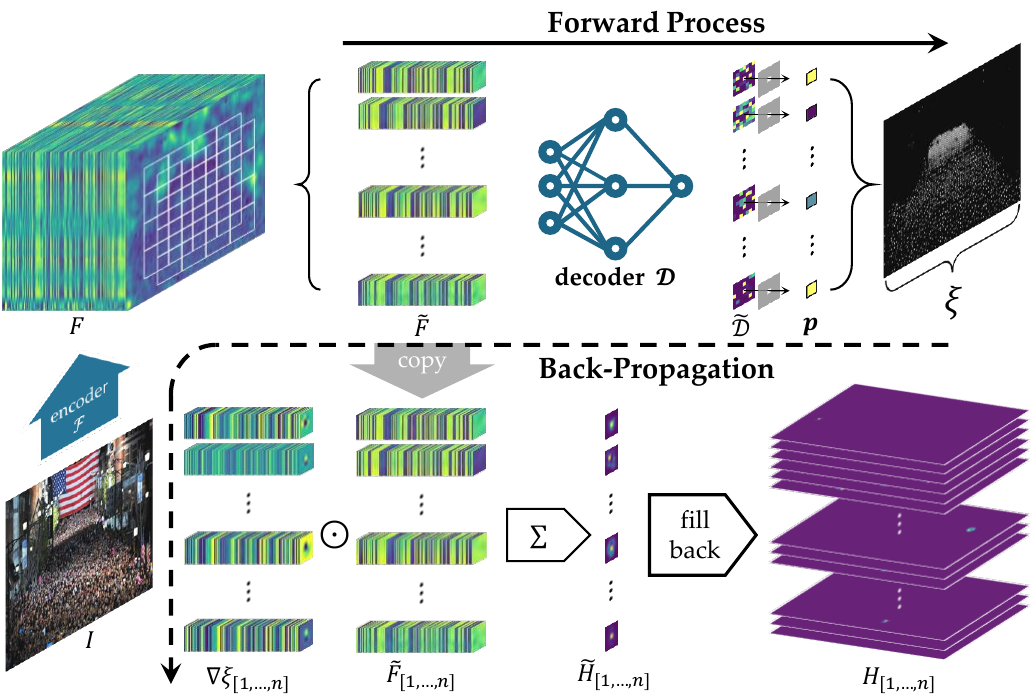}
	\caption{The generation process of PSAM.}
	\label{fig:psam_process}
\end{figure}

In a na\"{i}ve implementation, the computation of $\mbf{J}_\mac{D} \in \mathbb{R}^{n \times hwc}$ requires a  large memory and involves $n$ back-propagations through the computational graph, posing a challenge for visual explanation in pixel-level tasks. However, this issue can be resolved by computing gradients only between $\bsb{p}_{[q]}$ and features within its receptive field. Gradients in other areas are  zero, as there is no path connecting them in the computational graph. Thus, an efficient approach can be implemented, as illustrated in Fig.~\ref{fig:psam_process}.

In this implementation, the forward process begins with extracting sliding local blocks $\tilde{F}$ according to the receptive field $r$ of $\mac{D}$, \lwn{and then decoding $\tilde{F}$ through \eqref{eq:p2pnet} in parallel.} The final prediction $\bsb{p}$ is generated by stacking pixels located at the center of $r \times r$ patches denoted as in $\tilde{D}$:
\begin{gather}
	F \in \mathbb{R}^{h \times w \times c} 
	\rightarrow
	\tilde{F} \in \mathbb{R}^{(hw) \times (r \times r \times c)}, \label{eq:f} \\
	\mac{D}(\tilde{F}, \Theta_d) \rightarrow \tilde{D} \in \mathbb{R}^{n \times (r \times r)} \xrightarrow{\text{center of patches}} \bsb{p} \in \mathbb{R}^n. \label{eq:p2p_d2}
\end{gather}
The remaining $r^2-1$ pixels in each patch are discarded. 

Denoting the sum of the output as $\xi = \bsb{1}_n^\top \bsb{p}$, the gradient $\nabla \xi = \partial \xi / \partial \tilde{F}$ can be obtained via just one back-propagation, and the $q$-th gradient block, $\nabla \xi_{[q]}$, is equal to the gradient of $\bsb{p}_{[q]}$ with respect to $\tilde{F}_{[q]}$:
\begin{align}
	\nabla \xi_{[q]} = \tfrac{\partial \xi}{\partial \tilde{F}_{[q]}} = \tfrac{\partial \xi}{\partial \bsb{p}_{[q]}} \cdot \tfrac{\partial \bsb{p}_{[q]}}{\partial \tilde{F}_{[q]}} = \tfrac{\partial \bsb{p}_{[q]}}{\partial \tilde{F}_{[q]}}, \label{eq:gxqf}
\end{align}
since $\partial \xi / \partial \bsb{p}_{[q]} = 1$. After that, we can parallel compute the PSAM (denoted as $\tilde{H}_{[q]}$) within the receptive field of $\bsb{p}_{[q]}$, by substituting \eqref{eq:gxqf} into the formulation presented in \eqref{eq:hmap}:
\begin{equation}
	\tilde{H}_{[q]} = \max\big(\sum\nolimits_{k=1}^c \nabla\xi_{[q]} \odot \tilde{F}_{[q]}, 0\big). \label{eq:hq}
\end{equation}
As illustrated in the bottom of Fig.~\ref{fig:psam_process}, the complete PSAM $H_{[1,\dots, n]}$ in \eqref{eq:hmap} can be obtained by filling $\tilde{H}_{[1,\dots, n]}$ back into the corresponding regions of a fully zero matrix:
\begin{equation}
	H_{[q, \bsb{t}]} = 
	\begin{cases}
		\tilde{H}_{[q, \bsb{t}']}, & \text{if} ~~\bsb{t} \in \Omega_{q} \\
		0, & \text{otherwise}
	\end{cases}, \label{eq:hmap_f}
\end{equation}
where $\Omega_q$ represents the index set of $\bsb{p}_{[q]}$'s receptive field, and $\bsb{t}' \in \mathbb{R}^2$ is the transformed coordinate of $\bsb{t} \in \mathbb{R}^2$ in $H$, ensuring $F_{[\bsb{t}]} \equiv \tilde{F}_{[q, \bsb{t}']}$. In this way, the matrix memory required to obtain all $H_{[q]}$ is reduced from $\mathbb{R}^{n \times hwc}$ to $\mathbb{R}^{n \times r^2c}$ ($r^2 \ll hw$), and only one back-propagation is needed to obtain PSAMs for all pixels.

\subsection{Observation in PSAM}
\label{sec:psam}

\begin{figure}[t]
	\centering
	\includegraphics[width=0.47\textwidth]{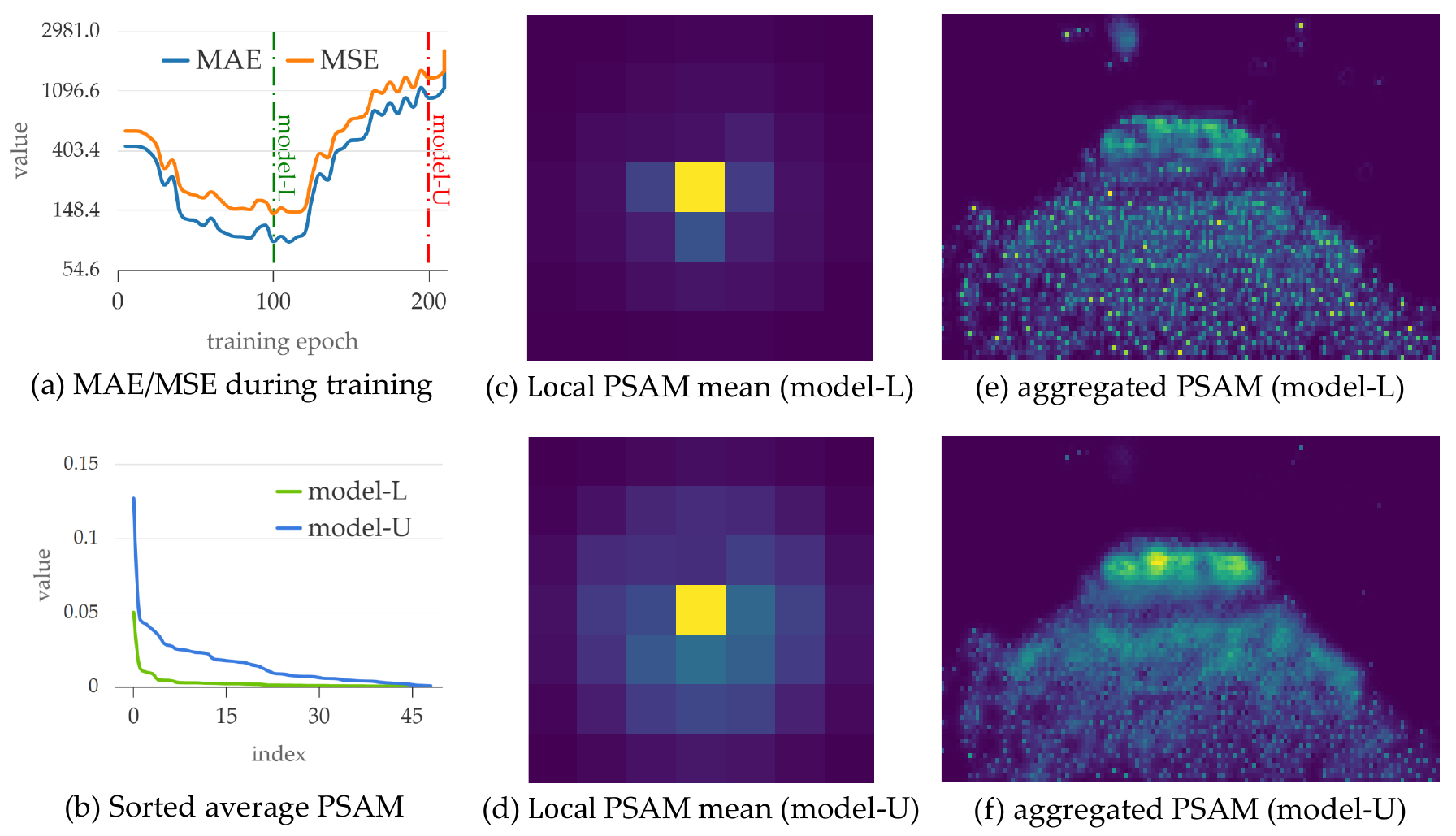}
	\caption{Observations in PSAM. (a) The training process, where model-L and model-U are extracted from the 100th and 200th epochs, respectively. (b) Comparing sorted values of PSAMs in the two models. (c) \& (d) Visualizing the average of local PSAM, and (e) \& (f) the aggregated PSAM to compare model-L and model-U from a global perspective.
	\NOTE{for (a), use semilogy to plot the MAE and MSE values, then it will be log plot.}\ANS{modified.}
	}
	\label{fig:psam_observation}
	\vspacex
\end{figure}

We next apply PSAM to visualize the model during semi-supervised counting.
The semi-supervised loss function combines \eqref{eq:bce} and \eqref{eq:bce_z}:
\begin{equation}
	\mac{L} = \alpha \mac{L}_{u} + (1 - \alpha)\mac{L}_l, \label{eq:loss}
\end{equation}
where $\alpha$ is a hyperparameter to balance the training of labeled and unlabeled data. Following \cite{psdlab}, we set $\alpha = 0$ during the first 100 epochs to generate valid pseudo labels. After that, $\alpha$ is gradually increased to 1. In Fig.~\ref{fig:psam_observation}(a), we plot how the 
MAE/MSE changes on the validation set during training. The training failed as \abc{evinced} by both metrics increasing significantly when unlabeled data were involved in training \abc{(after epoch 100)}. \abc{We denote `model-L' as the model at the 100-th epoch that is trained only on labeled data, and `model-U' as the model at the 200-th epoch that was trained with both labeled and unlabeled data.}

\abc{We next compare explanations of model-L and model-U using PSAM.} 
We visualize the concerned regions of the crowd by extracting the PSAMs corresponding to the foreground pixels ($\bsb{p}_{[q]} > 0.5$). In Fig.~\ref{fig:psam_observation}(c \& d) the element-wise mean of the PSAM patch ($\sum_{q}\tilde{H}_{[q]} \cdot (\bsb{p}_{[q]} > 0.5)$) is displayed to illustrate how the area of the concerned regions changes between model-L and model-U, while Fig.~\ref{fig:psam_observation}(b) presents a detailed value comparison using a line chart. On one hand, Fig.~\ref{fig:psam_observation}(b) shows that the values in model-U's PSAM are larger than those of model-L; on the other hand, Fig.~\ref{fig:psam_observation}(c \& d) shows that model-L's PSAM is more concentrated when compared to model-U. These observations reveal how \eqref{eq:bce_z} in P2P scheme affects the trained model: \emph{$\mac{L}_u$ leads to over-activation in the PSAM of foreground pixels, causing the decoder $\mac{D}$ to falsely identify surrounded pixels as foreground pixels as well}.

Next, we visualize the global PSAM of both models by compressing ${H}_{[q]}$ corresponding to foreground pixels: $\bar{H} = \sum\nolimits_{q=1}^n (\bsb{p}_{[q]} > 0.5) \cdot H_{[q]}$. As displayed in Fig.~\ref{fig:psam_observation}(e~\&~f), the aggregated PSAM of model-L is sparser than that of model-U, which is consistent with the observation for individual points presented in Fig.~\ref{fig:psam_observation}(c~\&~d). Another observation is that the background regions do not change significantly, indicating that almost all false positives occur in the neighborhood of the initial foreground pixels. This suggests that \emph{$\mac{L}_u$ does not alter the hyperplane separating crowd and non-crowd features in the high-dimensional space.}


\section{P2R Matching}

The observations described in last section reveal that $\mac{L}_u$ over-activates the values and enlarges the area of the PSAM corresponding to foreground pixels, which should be suppressed by the supervision of background part in \eqref{eq:bce}, but is actually zero in \eqref{eq:bce_z} ($\mac{L}_{u\texttt{[P2P]}}$ in \eqref{eq:bce_u}). \lws{This over-activation also confirms the nature of point annotations: \emph{rather than indicating the coordinates of a single pixel, each point in the GT or pseudo-label actually indicates a coarse local region that the corresponding pedestrian occupies.}}{This over-activation also confirms two important functions of the background item in \eqref{eq:bce}: it not only guides the model to classify the crowd and non-crowd regions in the image, but also \emph{suppresses the surrounding person region from being activated}. The latter is important since the neighboring pixels of a detected foreground pixel have similar features to the foreground pixel due to the CNN architecture, and thus may become activated if suppression is not applied.}
{This interpretation makes more sense because each person in a crowd image is represented by tens to thousands of pixels, not just one.} \lw{Supervising the region around the positive pixel with labels of 0 is crucial to a point-based counter.}
Therefore, during training, it would be more effective to adopt a point-to-region (P2R) matching strategy instead of a point-to-point (P2P) approach. Moreover, as suggested by the above interpretation, P2R is not limited in training with unlabeled data, but can also work in supervised learning with point supervision.

\subsection{Training with Labeled Data via P2R}
\label{sec:p2r_label}

\begin{figure}[t]
	\centering
	\includegraphics[width=0.46\textwidth]{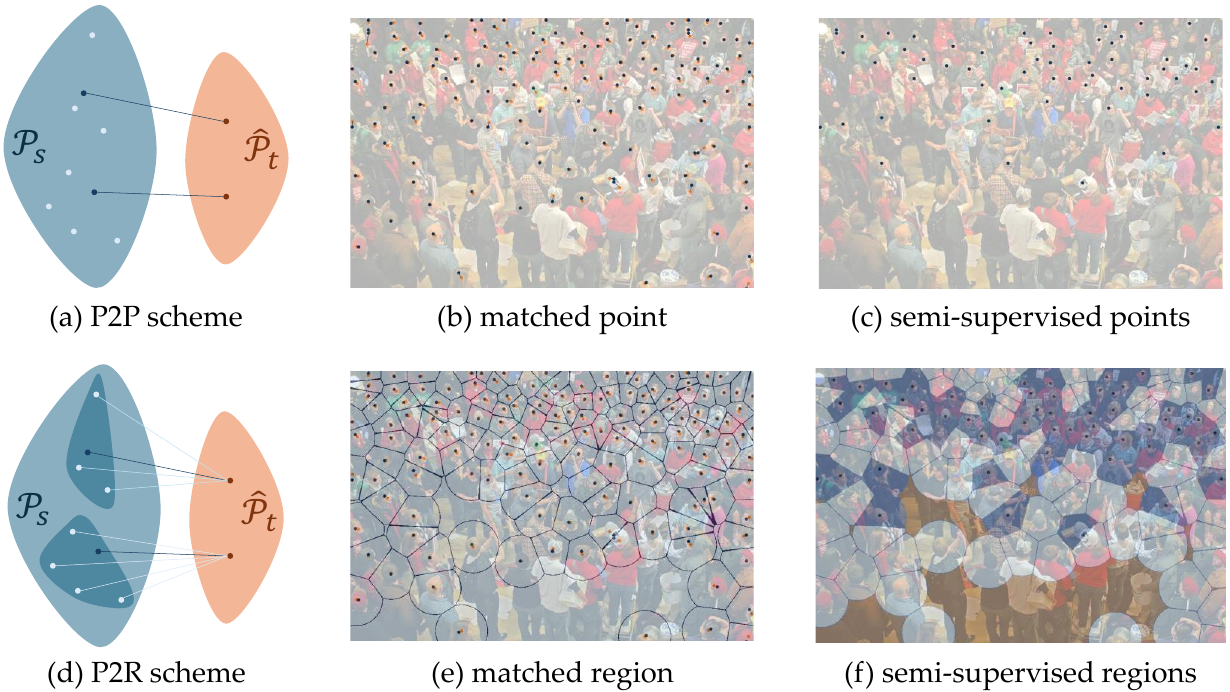}
	\caption{Difference between P2P and P2R matching. (a) and (d) demonstrate an overall difference. P2P focuses only on foreground pixels in $\mbf{P}_s$, while P2R considers background pixels as well. (b) and (e) show how the matching is performed in P2P and P2R, respectively. P2R segments out local regions for each pseudo-label, whereas P2P only detects one point. (c) and (f) illustrate how untrusted predictions are filtered. P2P retains only foreground pixels for loss computation, while P2R also keeps pixels in the neighborhoods of their corresponding pseudo points.}
	\label{fig:p2pr}
\end{figure}

Recalling the notation of the prediction $\mac{P} = \{ \bsb{p}, \bsb{x} \}$ in \eqref{eq:p2pout} and ground truth $\bsb{x}'$, the matching matrix $\mbf{M}$ in our P2R matching is the Hadamard product of two items:
\begin{align}
	\mbf{M} = \mbf{M}_f \odot (\bsb{\beta}^\top \mbf{1}_m)  \quad \in \mathbf{M}^{n\times m}. \label{eq:new_m}
\end{align}
The first item $\mbf{M}_f \in \{0, 1 \}^{n \times m}$ is a many-to-one matrix that assigns each pixel in $\mac{P}$ to its nearest point in $\bsb{x}'$:
\begin{align}
	{\mbf{M}_f}_{[ij]} = \begin{cases}
		1, & \text{if} ~~l_2(i, j) < l_2(i, k) ~~\forall k \neq j \\
		0, & \text{otherwise}
	\end{cases}, \label{eq:nearestn}
\end{align}
where $l_2(i, j) = \|\bsb{x}_{[i]} - {\bsb{x}'}_{[j]}\|_2$ is the $l_2$-distance between the $i$-th pixel's and the $j$-th point's coordinates. 
$\bsb{\beta} \in \{0, 1\}^{n}$ in the second item of \eqref{eq:new_m} is a vector marking foreground regions, by setting those pixels in $\mac{P}$ far from any points in $\bsb{x'}$ as zero:
\NOTE{you are using ``foreground'' to mean multiple things. In the previous section, foreground were pixels annotated as a person location. Now, ''foreground'' means the pixels close to an annotation. It's better to use consistent terminology, otherwise it will be confusing. }\ANS{here I use $\bsb{\beta}$ to denote the neighborhood of foreground pixel, instead of just foreground for clarity.}
\begin{align}
	\bsb{\beta}_{[i]} = \begin{cases}
		1, \quad \text{if} ~~\min\nolimits_j l_2(i, j) < \mu \\
		0, \quad \text{otherwise}
	\end{cases}, \label{eq:fg}
\end{align}
in which $\mu$ is a hyper parameter indicating the radius if the \lws{foreground}{neighborhood} of each GT point is considered as a circle.
As displayed in Fig.~\ref{fig:p2pr}(e), $\mbf{M}$ segments out a coarse local region that is consistent with the location of the GT point. After that, the learning objective $\hat{\bsb{p}}$ is derived by defining the cost matrix $\mbf{C} \in \mathbb{R}^{n \times m}$:
\begin{equation}
	\mbf{C}_{[ij]} = \begin{cases}
		\tau l_2(i, j) - \mac{S}(\bsb{p}_{[i]}), & \text{if} ~~ \mbf{M}_{[ij]} = 1 \\
		\infty, & \text{otherwise}
	\end{cases}. \label{eq:cost2}
\end{equation}
Then, the pixel with the minimum cost in each column of $\mbf{C}$ is marked as the potential foreground pixel:
\begin{align}
	\hat{\mbf{M}}_{[ij]} = \begin{cases}
		1, & \text{if} ~~ \mbf{C}_{[ij]} < \mbf{C}_{[kj]} ~~\forall k \neq i \\
		0, & \text{otherwise}
	\end{cases}. \label{eq:locmin}
\end{align}
Finally, the learning objective $\hat{\bsb{p}}$ is derived as $\hat{\bsb{p}} = \hat{\mbf{M}} \mbf{1}$, and the loss $\mac{L}_l$ is computed between $\bsb{p}$ and $\hat{\bsb{p}}$ according to BCE loss formulated as \eqref{eq:bce} during training with labeled data.

An extra benefit is that computing $\mbf{M}$ and $\hat{\mbf{M}}$ in P2R requires only finding the minimum of each row or column in the matrices, which is significantly faster than executing the Hungarian algorithm --- a time-consuming component of the P2P matching strategy~\cite{p2p}.

\subsection{Training with Unlabeled Data via P2R}

Recalling the pseudo-label $\mac{P}'_t = \{ \bsb{p}'_t, \bsb{x}'_t \}$ from the teacher and the supervised student output as $\mac{P}_s = \{ \bsb{p}_s, \bsb{x}_s \}$, the matching matrix $\mbf{M}_{st}$ between $\mac{P}_s$ and $\mac{P}'_t$ is computed via \eqref{eq:new_m} in P2R, and then combined with the confidence in \eqref{eq:conf} to mask unreliable pseudo labels. Additionally, pixels far from any pseudo points can be directly marked as reliable background pixels, as suggested by observations from the aggregated PSAM in Fig.~\ref{fig:psam_observation}(e \& f). Thus, the confidence diagonal matrix $\mbf{Z}$ is formulated as follows with $\bsb{\zeta} = \mathbbm{1}(\hat{\bsb{p}}'_t > \eta)$:
\begin{equation}
	\mbf{Z} = \diag\left[\mbf{M}_{st}\bsb{\zeta} + (\mbf{1}_n - \bsb{\beta}) \right] \label{eq:p2r_z}
\end{equation}
The final loss supervising unlabeled data $\mac{P}_s$ with pseudo labels is derived by substituting $\mbf{Z}$ in \eqref{eq:p2r_z} into \eqref{eq:bce_z}.

As displayed in Fig.~\ref{fig:p2pr}(f), the confidence for each pseudo point is propagated to it corresponding region. On one hand, reliable foreground parts (blue) contain a pixel whose pseudo-label is 1, while the pseudo-labels of its surrounding pixels are marked as 0. On the other hand, all pixels in the reliable background parts (khaki) are assigned pseudo-labels of 0 to train the counter with unlabeled data. The uncolored regions are ignored because their confidence is 0.

\lwn{The pseudo code and flow chart of the proposed P2R loss can be found in the supplemental material.}

\section{Experiments}

This section explores how the proposed P2R training approach enhances a point-based counting model using both labeled and unlabeled data across three parts: semi-supervised crowd counting, unsupervised domain adaptation (UDA) for crowd counting, and ablation studies. 

\lw{In the semi-supervised experiments, our method is implemented on four crowd counting datasets: ShTech A/B~\citep{mcnn}, UCF-QNRF~\cite{qnrf}, and JHU++~\cite{jhu}. Following DAC~\citep{dacl} and OT-M~\citep{otm}, three protocols are applied: 5\%, 10\%, and 40\% of labeled data, with the remaining crowd images involved in training without annotations. In the UDA part, labeled data from one domain, \textit{e.g.}, ShTech A, is used to initialize a counting model, while unlabeled data from another domain is utilized to capture the domain characteristics of the application scenes. In the ablation study, we test different hyperparameters to demonstrate their impact on the performance of P2R.}
%
%

\subsection{Semi-Supervised Counting}
\label{sec:ssc}

\NOTE{need a brief intro about the datasets and experiment protocol (which paper do you follow?). You can put a few sentences, and put details in the appendix. }
\ANS{I write it in the first version, but the page limitation does not allowed it.}

\begin{figure}
	\centering
	\includegraphics[width=0.48\textwidth]{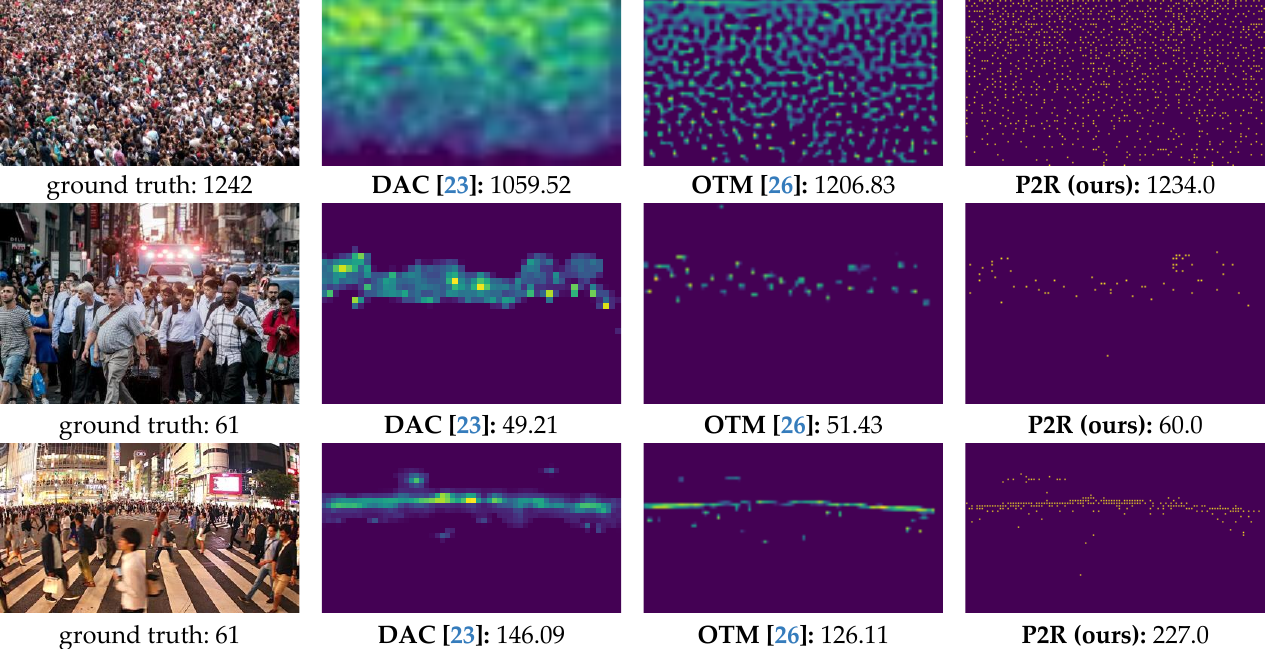}
	\caption{qualitative comparison with other models.}
	\label{fig:quvis}
\end{figure}

Tab.~\ref{tab:main_results} presents the experimental results of semi-supervised counting. Our method achieves the best performance in nearly all protocols across all crowd datasets. \lw{Besides, using 5\% data with P2R surpasses the fully-supervised learning with 10\% data and is even equivalent to other semi-supervised learning methods using 10\% data. Similar performances are also observed between the semi-supervised training with 10\% and 40\% labeled data. These results demonstrate that P2R is label-efficient and advantageous in semi-supervised learning.}
\NOTE{the below description is a bit dry. you can say that our method is label data efficient --- using 5\% data with ours is equivalent to 10\% data for other methods, etc. No need to mention all the numbers that can just be read from the table. }\ANS{revised.}
\lws{When the labeled data account for 5\%, P2R's MAE is smaller than 70 on ShTech \mbox{A~\cite{mcnn}}, which is a reduction of about 16\% compared to \mbox{OT-M~\cite{otm}}. When applied to ShTech B, the MAE of P2R is also better than the previous best result (9.1 \textit{vs.}~12.6). Additionally, outstanding performances are also achieved on \mbox{UCF-QNRF~\cite{qnrf}} and \mbox{JHU++~\cite{jhu}}, with MAEs of 100.1 and 77.8, respectively. When 10\% labeled data are provided, our P2R also achieves the lowest estimation errors on all datasets. Specifically, the MAEs in ShTech B, UCF-QNRF, and JHU++ are 8.4, 94.9, and 68.7, respectively, while the estimation errors in ShTech A are even close to those of \mbox{DAC~\cite{dac}} with 40\% annotated images (MAE: 65.2 \textit{vs.}~67.5; MSE: 114.6 \textit{vs.}~110.7). Furthermore, when the proportion of labeled data increases to 40\%, the performance is further improved and closer to that with 100\% labeled data when the model is a point-based counter. Although the MSE of P2R on the JHU++ dataset is 271.0, slightly worse than DAC (271.0 \textit{vs.}~260.0), it obtains a lower MAE than DAC (63.3 \textit{vs.}~65.1).}{}
\lwn{In Fig.~\ref{fig:quvis}, we visualize predictions of DAC~\citep{dacl}, OTM~\citep{otm}, and our P2R. The 3rd row is a failure case of P2R due to crowd motion blur. However, the overall performance of P2R is best.}

The last row of Tab.~\ref{tab:main_results} also compares  P2P and P2R under a fully-supervised learning scheme (100\% labels); P2R performs better than P2PNet on all crowd counting datasets. Additionally, as described in Section~\ref{sec:p2r_label}, P2R is faster than P2P due to the absence of the Hungarian algorithm. We compared the efficiency using an image with a resolution of $576 \times 960$ and 775 annotated points --- P2P requires an average of 0.4307 seconds for loss computation, while P2R only needs 0.0064 seconds, which is nearly 68 times faster  than P2P. These comparisons demonstrate that P2R surpasses P2P in both effectiveness and efficiency.

\begin{table}[t]
	\centering
	\resizebox{0.48\textwidth}{!}{
		\begin{tabular}{c|r|cc|cc|cc|cc}
			\toprule
			       {Label}         & \multirow{2}{*}{Methods} & \multicolumn{2}{c|}{ShTech A~\citep{mcnn}} & \multicolumn{2}{c|}{ShTech B~\citep{mcnn}} & \multicolumn{2}{c|}{UCF-QNRF~\citep{qnrf}} & \multicolumn{2}{c}{JHU++~\citep{jhu}} \\
			         Pct.          &                          &      MAE       &            MSE            &      MAE      &            MSE             &      MAE       &            MSE            &      MAE       &         MSE          \\ \midrule
			 \multirow{7}{*}{5\%}  &               label-only &      93.7      &           155.2           &     13.1      &            24.0            &     132.7      &           231.7           &     104.1      &        383.5         \\
			                       &             MT~\cite{mt} &     104.7      &           156.9           &     19.3      &            33.2            &     172.4      &           284.9           &     101.5      &        363.5         \\
			                       &           L2R~\cite{l2r} &     103.0      &           155.4           &     20.3      &            27.6            &     160.1      &           272.3           &     101.4      &        338.8         \\
			                       &             GP~\cite{gp} &     102.0      &           172.0           &     15.7      &            27.9            &     160.0      &           275.0           &      98.9      &        355.7         \\
			                       &           DAC~\cite{dac} &      85.4      &           134.5           &     12.6      &            22.8            &     120.2      &           209.3           &      82.2      &       {294.9}        \\
			                       &          OT-M~\cite{otm} &      83.7      &           133.3           &     12.6      &            21.5            &     118.4      &           195.4           &      82.7      &        304.5         \\ 
			                       &               P2R (ours) & \textbf{69.9}  &      \textbf{119.5}       & \textbf{9.1}  &       \textbf{16.6}        & \textbf{100.1} &      \textbf{182.5}       & \textbf{77.8}  &    \textbf{293.5}    \\ \midrule
			\multirow{8}{*}{10\%}  &               label-only &      84.0      &           138.3           &     10.7      &            19.2            &     112.4      &           186.8           &      78.7      &        305.5         \\
			                       &             MT~\cite{mt} &     319.3      &           94.5            &     15.6      &            24.5            &     156.1      &           245.5           &     250.3      &         90.2         \\
			                       &           L2R~\cite{l2r} &      90.3      &           153.5           &     15.6      &            24.4            &     148.9      &           249.8           &      87.5      &        315.3         \\
			                       &       IRAST~\cite{irast} &      86.9      &           148.9           &     14.7      &            22.9            &     135.6      &           233.4           &      86.7      &        303.4         \\
			                       &           DAC~\cite{dac} &      74.9      &           115.5           &     11.1      &            19.1            &     109.0      &           187.2           &      75.9      &        282.3         \\
			                       &          OT-M~\cite{otm} &      80.1      &           118.5           &     10.8      &            18.2            &     113.1      &           186.7           &      73.0      &        280.6         \\ 
			                       &            CU~\citep{cu} &      70.7      &           116.6           &      9.7      &            17.7            &     104.0      &          1644.2           &      74.8      &        281.6         \\
			                       &               P2R (ours) & \textbf{65.2}  &      \textbf{114.6}       & \textbf{8.4}  &       \textbf{14.5}        & \textbf{94.9}  &      \textbf{167.2}       & \textbf{68.7}  &    \textbf{272.3}    \\ \midrule
			\multirow{7}{*}{40\%}  &               label-only &      64.5      &           105.6           &      8.1      &            14.0            &      99.2      &           174.7           &      68.8      &        283.5         \\
			                       &             MT~\cite{mt} &      88.2      &           151.1           &     15.9      &            25.7            &     147.2      &           249.6           &     121.5      &        388.9         \\
			                       &           L2R~\cite{l2r} &      86.5      &           148.2           &     16.8      &            25.1            &     145.1      &           256.1           &     123.6      &        376.1         \\
			                       &           SUA~\cite{sua} &      68.5      &           121.9           &     14.1      &            20.6            &     130.3      &           226.3           &      80.7      &        290.8         \\
			                       &           DAC~\cite{dac} &      67.5      &           110.7           &      9.6      &            14.6            &      91.1      &           153.4           &      65.1      &    \textbf{260.0}    \\
			                       &          OT-M~\cite{otm} &      70.7      &           114.5           &      8.1      &            13.1            &     100.6      &           167.6           &      72.1      &        272.0         \\ 
			                       &               P2R (ours) & \textbf{55.6}  &       \textbf{95.0}       & \textbf{6.8}  &       \textbf{11.0}        & \textbf{86.0}  &      \textbf{144.3}       & \textbf{63.3}  &        271.1         \\ \midrule
			\multirow{2}{*}{100\%} &       P2PNet~\citep{p2p} &     52.74      &           85.06           &     6.25      &            9.90            &     85.32      &          154.50           &     61.25      &        258.65        \\
			                       &               P2R~(ours) & \textbf{51.02} &      \textbf{79.68}       & \textbf{6.17} &       \textbf{9.84}        & \textbf{83.26} &      \textbf{138.11}      & \textbf{58.83} &   \textbf{253.10}    \\ \bottomrule
		\end{tabular}
	}
	\caption{Comparison with other recent methods on four benchmark datasets under different labeled protocols.}
	\vspacex
	\label{tab:main_results}
\end{table}

\subsection{Unsupervised Domain Adaptation (UDA)}

UDA aims to transfer knowledge learned from a source domain to a target domain by designing modules or learning objectives that capture domain-agnostic features~\cite{mpc, c2mot, eadir, sfcn, ldaec, cvcs}. Unlike semi-supervised learning, where labeled and unlabeled data originate from the same domain, UDA typically involves a domain gap between the labeled (source domain) and unlabeled crowd data (target domain). This gap often manifests as differences in crowd distribution, density levels, and perspectives. Our P2R can also be applied to UDA. When an unlabeled image from the target domain is provided, the threshold automatically partitions its corresponding pseudo points into two categories: higher-score ($> \eta$) and lower-score ($< \eta$). The higher-score class is considered to be closer to the source domain, while the lower-score class is considered to be farther from the source domain. Thus, the higher-score pseudo points can participate in the training to gradually guide the model to learn to count in the target domain.

\begin{table}[t]
	\centering
	\resizebox{0.45\textwidth}{!}{
		\begin{tabular}{c|r|cc|cc|cc|cc}
			\toprule
			\multicolumn{2}{c|}{\multirow{2}{*}{METHODS}}            & \multicolumn{2}{c|}{A $\rightarrow$ B} & \multicolumn{2}{c|}{A $\rightarrow$ Q} & \multicolumn{2}{c|}{B $\rightarrow$ A} & \multicolumn{2}{c}{B $\rightarrow$ Q} \\
			\multicolumn{1}{c}{}&                                           &      MAE      &          MSE           &      MAE       &          MSE          &      MAE      &          MSE           &      MAE       &         MSE          \\ \midrule
			\multirow{2}{*}{\scriptsize \textit{DG}}  & DCCUS~\citep{dccus} &     12.6      &          24.6          &     119.4      &         216.6         &     121.8     &         203.1          &     179.1      &        316.2         \\
			&      MPCount~\citep{mpc}                                     &    {11.4}     &         {19.7}         &     115.7      &        {199.8}        &  \udl{99.6}   &      \udl{182.9}       &  \udl{165.6}   &     \udl{290.4}      \\ \midrule
			\multirow{5}{*}{\scriptsize \textit{UDA}} & BL~\citep{bl} (w/o DA) &     15.9      &          25.8          &     166.7      &         287.6         &     138.1     &         228.1          &     226.4      &        411.0         \\
			&  RBT~\citep{rbt}                                         &     13.4      &          29.3          &     175.0      &         294.8         &     112.2     &         218.2          &     211.3      &        381.9         \\
			& C$^2$MoT~\citep{c2mot}                                          &     12.4      &          21.1          &     125.7      &         218.3         &     120.7     &         192.0          &     198.9      &        368.0         \\
			& FGFD~\citep{fgfd}                                          &     12.7      &          23.3          &     124.1      &         242.0         &     123.5     &         210.7          &     209.7      &        384.7         \\
			& FSIM~\citep{fsim}                                          &  \udl{11.1}   &       \udl{19.3}       &     105.3      &    \textbf{ 191.1}    &     120.3     &         202.6          &     194.9      &        324.5         \\ \midrule
			\multirow{2}{*}{\scriptsize \textit{UDA}} & P2R (w/o DA) &     25.6      &          35.7          &     155.1      &         286.8         &     130.2     &         229.1          &     173.3      &        329.8         \\
			&  P2R (w/~~ DA)                                         & \textbf{10.6} &     \textbf{18.7}      & \textbf{105.3} &      \udl{194.8}      & \textbf{89.3} &     \textbf{176.0}     & \textbf{139.5} &    \textbf{243.1}    \\ \bottomrule
		\end{tabular}}
	\caption{Unsupervised domain adaptation for crowd counting. The LHS and RHS of ``$\rightarrow$'' represent adaptation direction. Q, A, and B indicate UCF-QNRF~\cite{qnrf}, ShTech A, and ShTech B~\cite{mcnn}. \lwn{\textit{DG} indicates domain generalization. The training of UDA requires samples from the target domain, whereas DG does not.}}
	\vspacex
	\label{tab:uda}
\end{table}

We conduct experiments on four protocols \lw{following \cite{c2mot}}.
 The results are presented in Tab.~\ref{tab:uda}. \lw{The last two rows demonstrate the results of the proposed P2R. The poor performance obtained without domain adaptation highlights the domain gap between the source data and target data. However, when the unlabeled data from the target domain is involved in training, the estimation errors are significantly reduced, and the results surpass previous methods with complex network structures for domain adaptation. These advanced experimental results demonstrate that the threshold works as expected to capture samples close to the source domain and gradually teaches the model to count in the target domain.}

\subsection{Ablation Study}

\begin{figure}[t]
	\centering
	\begin{tabular}{cc}
		\includegraphics[width=0.205\textwidth]{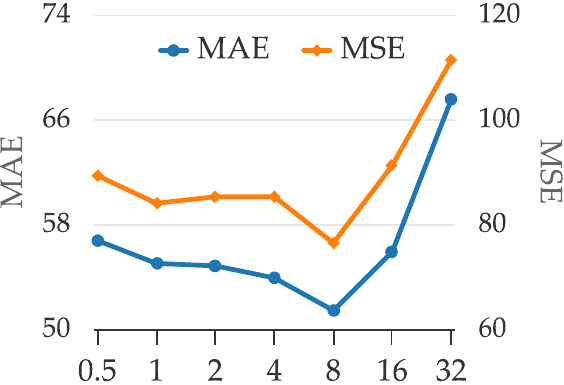} &
		\includegraphics[width=0.205\textwidth]{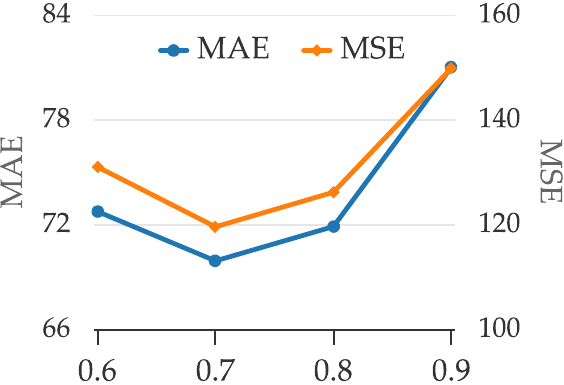} \\
		(a) errors \textit{vs.} $\tau$ & (b) errors \textit{vs.} $\eta$ \\
		\includegraphics[width=0.205\textwidth]{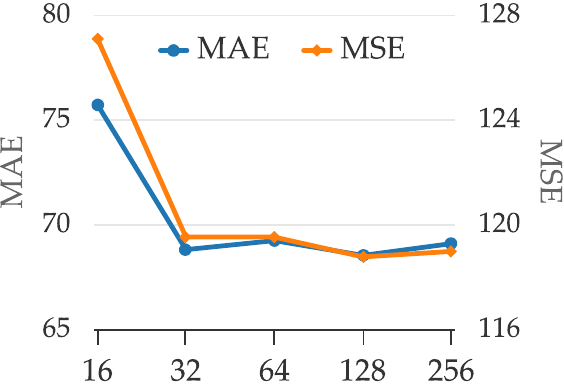} &
		\includegraphics[width=0.205\textwidth]{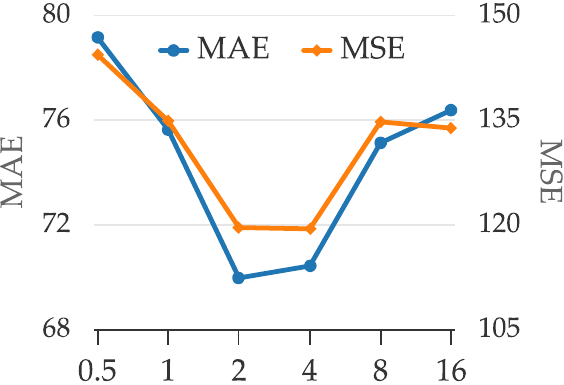} \\
		 (c) errors \textit{vs.} $\mu$ & (d) errors \textit{vs.} $\frac {\alpha }{1 - \alpha}$
	\end{tabular}
	\caption{The estimation errors vs. various hyperparameters. (a) is conducted on the fully-supervised ShTech A since it is irrelevant to semi-supervised counting, while (b)-(d) are conducted with 5\% labeled data in ShTech A.}
	\label{fig:ablation}
	\vspace{-5mm}
\end{figure}

We next conduct ablation studies to investigate the influence of hyperparameters in P2R, including  $\tau$ in \eqref{eq:cost2}, the threshold $\eta$, $\mu$ in \eqref{eq:fg}, and $\alpha$ in \eqref{eq:loss}. The results are  in Fig.~\ref{fig:ablation}

{\bf The impact of $\tau$ in \eqref{eq:cost2}.} $\tau$ is a hyperparameter to balance the distance and foreground score when selecting the foreground pixel from the matched region. When $\tau \rightarrow \infty$, the predicted score is ignored, which is equivalent to computing cross-entropy between the prediction and the ground truth point map. If $\tau = 0$, the pixel with the highest score is chosen as the foreground in $\hat{\bsb{p}}$. However, Fig.~\ref{fig:ablation}(a) shows that neither extreme is optimal. The best performance is achieved when $\tau = 8$.

{\bf The effect of threshold $\eta$.} $\eta$ defines the threshold to distinguish reliable and unreliable regions. Since all pseudo points are selected by picking pixels whose score is greater than 0.5 in the prediction, the threshold should also be greater than 0.5. Moreover, a lower $\eta$ introduces unreliable pseudo-labels into training, while a higher $\eta$ cannot fully utilize the unlabeled data as most parts cannot participate in training. As shown in Fig.~\ref{fig:ablation}(b), the estimation errors exhibit a trend of first decreasing and then increasing within the range of 0.6 to 0.9, with the lowest errors occurring when the threshold is set to 0.7.

{\bf The radius of each point $\mu$.} The vector $\bsb{\beta}$ in \eqref{eq:fg} derived from $\mu$ marks the foreground points' neighborhood and background regions. It not only restricts the positions where foreground pixels appear by substituting $\bsb{\beta}$ into \eqref{eq:new_m}, but also directly identifies reliable background parts via the second item in \eqref{eq:p2r_z} based on the observation of PSAM in Section~\ref{sec:psam}. Fig.~\ref{fig:ablation}(c) presents how $\mu$ affects the counting performance. 
It shows the MAE remains around 68 when $\mu$ increases from 32 to 256, indicating that a larger $\mu$ does not significantly affect the results. This is consistent with the observation in PSAM since pixels far from the foreground points are less likely to be activated during training.

{\bf The weight of unlabeled loss.} Fig.~\ref{fig:ablation}(d) presents how $\alpha$ in \eqref{eq:loss} affects the performance. A large $\alpha$ focuses more on unlabeled data, leading to training failure, while a small $\alpha$ may cause the model to overfit on labeled data. \lwn{In the experiments with ShTech A and 5\% labeled data, $\frac{\alpha}{1 - \alpha} = 2$, \textit{i.e.}, $\alpha = 2/3$, results in the best performance.} Additionally, as long as $\alpha$ is not 0, the point-based counter can benefit from unlabeled data, as the maximum MAE/MSE in Fig.~\ref{fig:ablation}(d) are smaller than 80/150, which is much better than the model trained with only 5\% labeled data (MAE: 93.7, MSE: 155.2).

\begin{figure}
	\centering
	\includegraphics[width=0.45\textwidth]{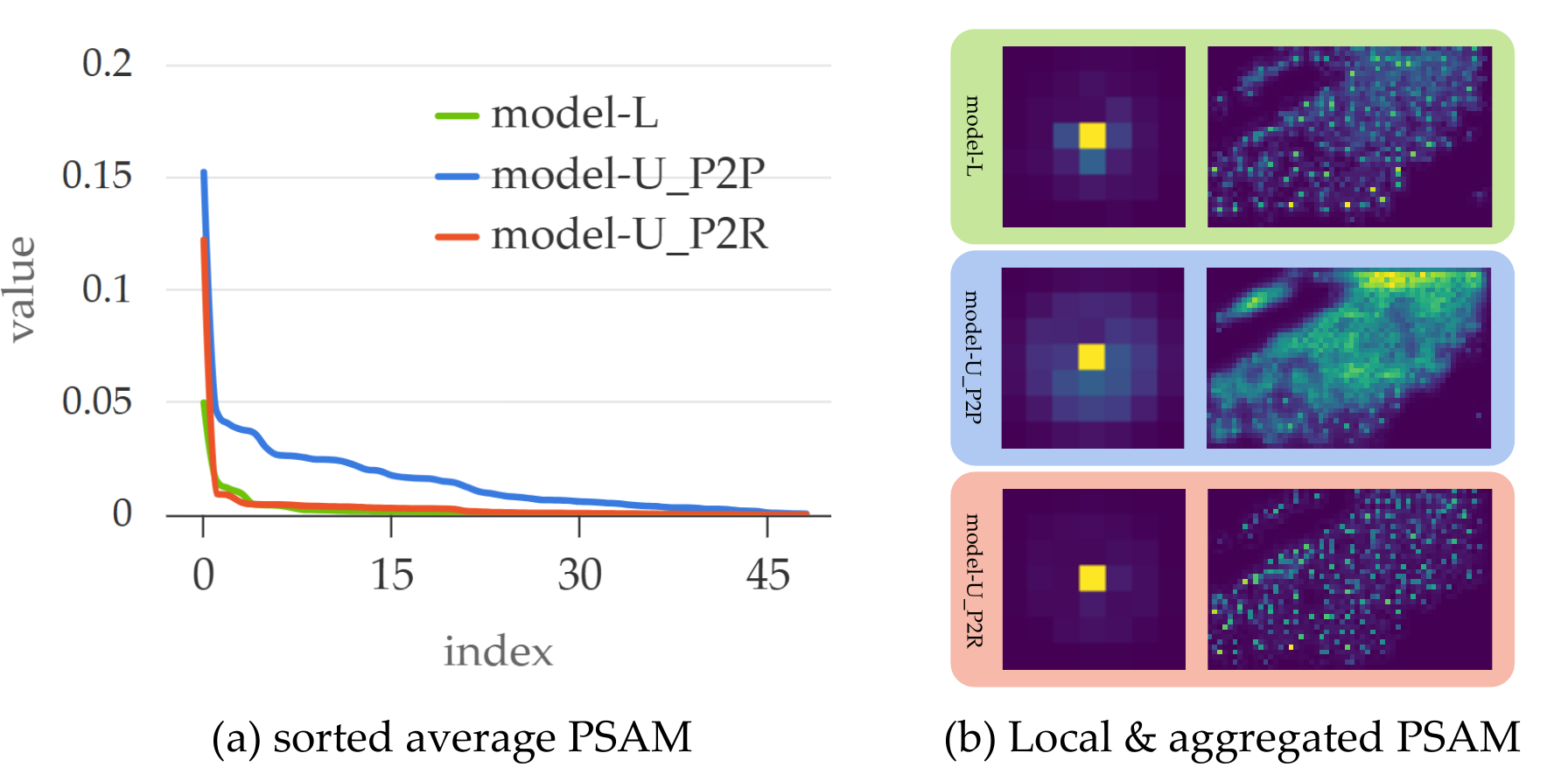}
	\caption{\lwn{The comparison of PSAMs among different training schemes. Model-L (green) is trained with only labeled data; model-U\_P2P (blue) is trained with both labeled data and unlabeled data using the P2P scheme, while model-U\_P2R (orange) is trained with the proposed P2R scheme.}}
	\label{fig:psam2}
	\vspace{-4mm}
\end{figure}

{\bf PSAM comparison.} Fig.~\ref{fig:psam2} visualizes the PSAMs of models trained with the P2P and P2R schemes. Comparing model-L (green) and model-U\_P2R (blue), the latter effectively restricts the PSAM to the neighborhood of the concerned foreground pixel, and its aggregated PSAM is also sparser than that of model-L, demonstrating the effectiveness of P2R. Comparing model-U\_P2P and model-U\_P2R in Fig.~\ref{fig:psam2}(a), the values of the PSAM are similar at the non-peak indexes, but the latter has a higher activation at the peak position compared to the former, indicating that semi-supervised learning further enhances the identification ability of the point-based counter.

\section{Conclusion}

This paper presents a P2R scheme to train a point-based counter for semi-supervised crowd counting. At the beginning, we establish a pseudo-labeling framework based on P2P matching, but its formulation is ill-posed since the confidence is only applicable to foreground pixels and cannot be propagated to backgrounds. To observe how the foreground and background feature are learned, we propose PSAM to  visually interpret the concerned region and activation value of all foreground pixels. From the visualized activation map, it is observed that the neighborhood of each foreground pixel is over-activated and falsely recognized as instances by the decoder. Based on this observation, we replace P2P with P2R by segmenting the prediction into multiple regions containing the corresponding instances. Thus, pixels in the neighborhood of the concerned pseudo points can share the weights with the corresponding pixels. Additionally, the time-consuming Hungarian algorithm is no longer necessary in P2R. Several outstanding experimental results in semi-supervised counting and unsupervised domain adaptation demonstrate the advantages of our P2R matching strategy.

\vspace{-2mm}
\paragraph{Acknowledgements.} \lwn{This work was supported by a grant from City University of Hong Kong (Project No. 7030010).}

{
    \small
    \bibliographystyle{ieeenat_fullname}
    \bibliography{main}
}

  \newpage
\maketitlesupplementary
\newcommand{\sadd}[1]{\textcolor{magenta}{#1}}

\begin{algorithm}[t]
	\newcommand\mycommfont[1]{\scriptsize\ttfamily\textcolor{blue}{#1}}
	\SetCommentSty{mycommfont}
	\SetKwInput{KwInput}{Input}                
	\SetKwInput{KwOutput}{Output}              
	\SetKwInput{KwHypara}{Hyperparameter}              
	
	\DontPrintSemicolon
	\KwHypara{$\mu, \tau$, and $\lambda$}
	\tcc{ Score map predicted by model:}
	\KwInput{ $\mac{P} = \{\bsb{p} \in \mathbb{R}^{n}, \bsb{x} \in \mathbb{R}^{n\times 2}\}$}
	\tcc{The coordinates of ground truth points:}
	\KwInput{$\bsb{x}' \in \mathbb{R}^{m\times 2}$}
	\KwOutput{loss value $\mac{L}_l$.}
	\tcc{ Compute the $l_2$ distance matrix $\mbf{L}_2$}
	$\mbf{L}_2 \in \mathbb{R}^{n\times m}:~~ {\mbf{L}_2}_{[i, ~j]} \gets \|\bsb{x}_{[i]} - {\bsb{x}'}_{[j]}\|_2$
	
	\tcc{Obtain the minimum $\bsb{d}$ and the corresponding column index $\bsb{k}$ of each row in $\mbf{L}_2$}
	Initialize $\bsb{d} \in \mathbb{R}^n$ and $\bsb{k} \in \mathbb{N}^{n}$
	
	\For {$i \gets 1$ to $n$} {
		$\bsb{d}_{[i]} \gets \min\nolimits_j {\mbf{L}_2}_{[i,~j]}$
		
		$\bsb{k}_{[i]} \gets \mathop{\arg\min}\nolimits_{j} {\mbf{L}_2}_{[i,~j]}$
	}
	
	\tcc{Compute the P2R matching matrix $\mbf{M}$ in \eqref{eq:new_m}}
	$\mbf{M}_f \gets \mbf{0}_{n\times m}, \quad \bsb{\beta} \gets \mbf{0}_n$
	
	\For {$i \gets 1$ to $n$} {
		${\mbf{M}_f}_{\left[i, ~\bsb{k}_{[i]}\right]} \gets 1, \quad \bsb{\beta}_{[i]} \gets (\bsb{d}_{[i]} < \mu)$
		
	}
	$\mbf{M} \gets \mbf{M}_f \odot (\bsb{\beta}^\top\mbf{1}_m)$
	
	\tcc{Compute the cost $\mbf{C}$ in \eqref{eq:cost2} for foreground pixel selection, in which the subsection in \eqref{eq:cost2} can be implemented by element-wise division}
	$\mbf{C} \in \mathbb{R}^{n\times m}$:~~$\mbf{C}_{[i, ~j]} \gets \tau{\mbf{L}_2}_{[i, ~j]} + \log \left(\frac 1 {\bsb{p}_{[i]}} - 1\right)$
	
	$\mbf{C} \gets \mbf{C} \odiv {\mbf{M}}$
	
	\tcc{Estimate the learning objective $\hat{\bsb{p}}$ by marking the minimum in each column of $\mbf{C}$ using $\hat{\mbf{M}}$}
	
	$\hat{\mbf{M}} \gets \mbf{0}_{n\times m}$
	\For {$j \gets 1$ to $m$} {
		$k \gets \mathop{\arg\min}_i \mathbf{C}_{[i, ~j]}$ \\
		$\hat{\mbf{M}}_{[k, ~j]} \gets 1$
	}
	$\hat{\bsb{p}} \gets \hat{\mbf{M}}^\top\mbf{1}_m$
	
	\tcc{compute the weighted binary cross entropy}
	$\mac{L}_{l} \gets - \lambda \hat{\bsb{p}}^\top \log \bsb{p} - (\mbf{1}_n - \hat{\bsb{p}})^\top \log (\mbf{1}_n - \bsb{p})$
	
	\Return{$\mac{L}_{l}$}
	\caption{P2R in fully-supervised training}
	\label{algo:f-sup}
\end{algorithm}

\begin{algorithm}[t]
	\newcommand\mycommfont[1]{\scriptsize\ttfamily\textcolor{blue}{#1}}
	\SetCommentSty{mycommfont}
	\SetKwInput{KwInput}{Input}                
	\SetKwInput{KwOutput}{Output}              
	\SetKwInput{KwHypara}{Hyperparameter}              
	
	\DontPrintSemicolon
	\KwHypara{$\mu, \tau$, \sadd{$\eta$,} and $\lambda$}
	\tcc{\sadd{prediction of student:}}
	\KwInput{ $\mac{P}_s = \{\bsb{p} \in \mathbb{R}^{n}, \bsb{x} \in \mathbb{R}^{n\times 2}\}$}  
	\tcc{\sadd{pseudo-labels generated by teacher:}}
	\KwIn{\sadd{$\mac{P}'_t = \{\bsb{p}'_t \in \mathbb{R}^m, \bsb{x}'_t \in \mathbb{R}^{m\times 2}\}$}}
	\KwOutput{loss value \sadd{$\mac{L}_u$}.}
	\tcc{ Compute the $l_2$ distance matrix $\mbf{L}_2$}
	$\mbf{L}_2 \in \mathbb{R}^{n\times m}:~~ {\mbf{L}_2}_{[i, ~j]} \gets \|\sadd{{\bsb{x}_s}_{[i]} - {\bsb{x}'_t}_{[j]}}\|_2$
	
	\tcc{Obtain the minimum $\bsb{d}$ and the corresponding column index $\bsb{k}$ of each row in $\mbf{L}_2$}
	Initialize $\bsb{d} \in \mathbb{R}^n$ and $\bsb{k} \in \mathbb{N}^{n}$
	
	\For {$i \gets 1$ to $n$} {
		$\bsb{d}_{[i]} \gets \min\nolimits_j {\mbf{L}_2}_{[i,~j]}$
		
		$\bsb{k}_{[i]} \gets \mathop{\arg\min}\nolimits_{j} {\mbf{L}_2}_{[i,~j]}$
	}
	
	\tcc{Compute the P2R matching matrix $\mbf{M}_{st}$}
	$\mbf{M}_f \gets \mbf{0}_{n\times m}, \quad \bsb{\beta} \gets \mbf{0}_n$
	
	\For {$i \gets 1$ to $n$} {
		${\mbf{M}_f}_{\left[i, ~\bsb{k}_{[i]}\right]} \gets 1, \quad \bsb{\beta}_{[i]} \gets (\bsb{d}_{[i]} < \mu)$
		
	}
	$\sadd{\mbf{M}_{st}} \gets \mbf{M}_f \odot (\bsb{\beta}^\top\mbf{1}_m)$
	
	\tcc{Compute the cost $\mbf{C}$ in \eqref{eq:cost2} for foreground pixel selection, in which the subsection in \eqref{eq:cost2} can be implemented by element-wise division}
	$\mbf{C} \in \mathbb{R}^{n\times m}$:~~$\mbf{C}_{[i, ~j]} \gets \tau{\mbf{L}_2}_{[i, ~j]} + \log \left(\frac 1 {\bsb{p}_{[i]}} - 1\right)$
	
	$\mbf{C} \gets \mbf{C} \odiv \sadd{\mbf{M}_{st}}$
	
	\tcc{Estimate the learning objective $\hat{\bsb{p}}$ by marking the minimum in each column of $\mbf{C}$ using $\hat{\mbf{M}}$}
	
	$\hat{\mbf{M}} \gets \mbf{0}_{n\times m}$
	\For {$j \gets 1$ to $m$} {
		$k \gets \mathop{\arg\min}_i \mathbf{C}_{[i, ~j]}$ \\
		$\hat{\mbf{M}}_{[k, ~j]} \gets 1$
	}
	$\hat{\bsb{p}} \gets \hat{\mbf{M}}^\top\mbf{1}_m$
	
	\tcc{\sadd{Compute the confidence diagonal matrix $\mbf{Z}$}}
	
	\sadd{$\bsb{\xi} \in \{0, 1\}^{m}:~~ \bsb{\xi}_{[i]} \gets {\bsb{p}'_t}_{[i]} > \eta$}
	
	\sadd{$\mbf{Z} \gets \mbf{0}_{n\times n}$}
	
	\For {\sadd{$i \gets 1$ to n}} {
		\sadd{$\mbf{Z}_{[i, ~i]} \gets {\mbf{M}_{st}}_{[i, ~:]}^\top\bsb{\xi} + (1 - \bsb{\beta}_{[i]})$}
	}
	
	\tcc{compute the weighted binary cross entropy}
	$\sadd{\mac{L}_{u}} \gets - \lambda \hat{\bsb{p}}^\top\sadd{\mbf{Z}}\log \bsb{p} - (\mbf{1}_n - \hat{\bsb{p}})^\top\sadd{\mbf{Z}}\log (\mbf{1}_n - \bsb{p})$
	
	\Return{\sadd{$\mac{L}_{u}$}}
	\caption{P2R in semi-supervised training}
	\label{algo:s-sup}
\end{algorithm}

In this supplemental material, we present the following contents:
\begin{compactitem}
	\item Sec.~\ref{sec:pcode}: Pseudo-code of P2R for training.
	\item Sec.~\ref{sec:imd}: Implementation details of P2R.
	\item Sec.~\ref{sec:si}: Ablation study on $\mac{S}(\cdot)$ \textit{vs.} $I(\cdot)$ in \eqref{eq:cost}.
	\item Sec.~\ref{sec:proof}: Proof of the unlabeled loss \eqref{eq:bce_u}.
	\item Sec.~\ref{sec:other}: Comparison of P2R with state-of-the-art loss functions in fully-supervised crowd counting.
	\item Sec.~\ref{sec:vis}: Visualization of P2R predictions on UCF-QNRF dataset images.
\end{compactitem}

\section{Pseudo Code of P2R}
\label{sec:pcode}

In Algo.~\ref{algo:f-sup} and Algo.~\ref{algo:s-sup}, we present the pseudo-code to compute the loss for data with ground truth (GT) labels and pseudo-labels, respectively. In Algo.~\ref{algo:s-sup}, the parts that differ from those in Algo.~\ref{algo:f-sup} are highlighted in \sadd{purple}. The comparison between these two algorithms shows that most of the steps are the same, except for the involvement of confidence computation in Algo.~\ref{algo:s-sup}.

Also note that the computation of P2R is efficient since the loops in Algorithm~\ref{algo:f-sup} and Algorithm~\ref{algo:s-sup} can be written as matrix operations, which can be executed in parallel on a GPU.

\section{Implementation Details}
\label{sec:imd}

{\bf Data pre-processing.} Images in the datasets are cropped into $256 \times 256$ for training. For labeled images, we apply horizontal flips to each cropped sample with a probability of 0.5 and randomly resize the images with a scale factor between 0.7 and 1.3. For unlabeled data, the set of weak data augmentation operations is the same as those for labeled data, while strong data augmentation includes adjustments of brightness, contrast, saturation, and hue, conversion from color images to grayscale images, the addition of Gaussian blur with different kernel sizes, and cutout. Besides, Cutout is also implemented, as displayed in Fig.~\ref{fig:fixmatch}.

{\bf Training process.} In all experiments, we train the model for 1500 epochs with a batch size of 16. Only labeled data are used in the first 100 epochs for initialization. After that, $\alpha$ in \eqref{eq:loss} is gradually increased from 0 to 1 with a step of 0.01. Adam~\cite{adam} serves as the optimizer, with a learning rate of $5 \times 10^{-5}$ for the decoder $\mac{D}$ and $1 \times 10^{-5}$ for the backbone $\mac{F}$. Furthermore, the loss in the cut-out patch is directly set to 0 to avoid unreasonable optimization.

\section{Ablation Study on Cost Function}
\label{sec:si}

In \eqref{eq:cost}, we use the inverse sigmoid function,
\begin{equation}
	\mac{S}(p) = -\log \left(\frac{1}{p} - 1\right),
\end{equation}
rather than the identity operator used in vanilla P2PNet~\cite{p2p} for better performance. We present the empirical results in Table~\ref{tab:abs} to demonstrate its advantage. It shows that the inverse sigmoid function performs better than the identity operator on both P2P~\cite{p2p} and P2R.
 
 \begin{table}
 	\centering
 	\begin{tabular}{r|cc|cc}
 		\toprule
 		\multirow{2}{*}{Method} & \multicolumn{2}{c|}{Identity} & \multicolumn{2}{c}{Inv-Sigmoid} \\
 		    & MAE & MSE & MAE & MSE \\
 		\midrule
 		P2P~\citep{p2p} & 52.74 & 85.60 & \textbf{52.50} & \textbf{82.94} \\
 		P2R~(ours) & 53.30 & 83.01 & \textbf{51.02} & \textbf{79.68} \\
 		\bottomrule
 	\end{tabular}
 	\caption{Ablation study on $\mac{S}(\cdot)$.}
 	\label{tab:abs}
 \end{table}

\section{The Proof of Ill-Posed Unlabeled Loss}
\label{sec:proof}
Under the P2P framework, we demonstrate that the loss function for unlabeled data, formulated as \eqref{eq:bce_z}, is ill-posed, since the second term for the background part is set to 0, as shown in \eqref{eq:bce_u}. This is easy to follow because
\begin{align}
	\mbf{Z}\hat{\bsb{p}}_t &= \diag(\mbf{M}_{st}\bsb{\zeta})(\mbf{M}_{st}\mbf{1}_n) \label{eq:z0}  \\
	&= (\mbf{M}_{st}\bsb{\zeta}) \odot (\mbf{M}_{st}\mbf{1}_n) = \mbf{M}_{st}\bsb{\zeta}. \label{eq:z1} 
\end{align}
Equation in \eqref{eq:z1} holds since each row of $\mbf{M}_{st}$ is an all-zero or one-hot vector \lwn{since it is matrix result of bipartite-graphs matching between prediction and GT.} Taking any row of $\mbf{M}_{st}$ and denoting it as $\bsb{m} \in \mathbb{R}^m$, we have:
\begin{align}
	\bsb{m}^\top\bsb{\zeta} &= \begin{cases}
		1, &\text{if} ~~ \exists j: \bsb{m}_{[j]} = \bsb{\zeta}_{[j]} = 1 \\
		0, &\text{otherwise}
	\end{cases}, \label{eq:mz} \\
	\bsb{m}^\top\mbf{1}_m &= \begin{cases}
		1, &\text{if} ~~ \exists j: \bsb{m}_{[j]} = 1 \\
		0, &\text{otherwise}
	\end{cases}, \label{eq:m1}
\end{align}
which can be combined to result in:
\begin{align}
	(\bsb{m}^\top\bsb{\zeta})(\bsb{m}^\top\mbf{1}_m) &= \begin{cases}
		1, &\text{if} ~~ \exists j: \bsb{m}_{[j]} = \bsb{\zeta}_{[j]} = 1 \\
		0, &\text{otherwise}
	\end{cases}. \label{eq:mz1}
\end{align}
Note that \eqref{eq:mz} and \eqref{eq:mz1} have the same formulation, thus the following equation holds:
\begin{align}
	(\bsb{m}^\top\bsb{\zeta}) (\bsb{m}^\top\mbf{1}_m) &= \bsb{m}^\top\bsb{\zeta} \\
	\Rightarrow \quad (\mbf{M}_{st}\bsb{\zeta}) \odot (\mbf{M}_{st}\mbf{1}_n) &= \mbf{M}_{st}\bsb{\zeta}.
\end{align}
Due to the equality of \eqref{eq:z0} and \eqref{eq:z1}, the following relationship can be derived:
\begin{align}
	\mbf{Z}\hat{\bsb{p}}_t = \mbf{M}_{st}\bsb{\zeta} &= \diag(\mbf{M}_{st}\bsb{\zeta})\mbf{1}_n = \mbf{Z}\mbf{1}_n,
\end{align}
which shows that the second term in \eqref{eq:bce_z} is 0:
\begin{align}
	\mbf{Z}(\mbf{1}_n - \hat{\mbf{p}}_t) &= \mbf{0}_n \\
	\Rightarrow \quad (\mbf{1}_n - \hat{\mbf{p}}_t)^\top\mbf{Z}\log(\mbf{1}_n - \bsb{p}_s) &= 0.
\end{align}

\section{Comparison with Other Losses}
\label{sec:other}
\begin{table}[t]
	\centering
	\scalebox{0.77}{\begin{tabular}{rc|cc|cc|c}
			\hline
			\multicolumn{1}{c}{Loss} & Point-based & \multicolumn{2}{c|}{ShTech B} & \multicolumn{2}{c|}{QNRF} & \multirow{2}{*}{FPS} \\
			\multicolumn{1}{c}{function} & counting model & MAE & MSE & MAE & MSE & \\
			\hline
			L2~\citep{mcnn} & \ding{55} & 7.6 & 13.0 & 102.0 & 171.4 & \textbf{1503.8} \\
			BL~\cite{bl} & \ding{55} & 7.7 & 12.7 & 88.7 & 154.8 & 274.22 \\
			GL~\citep{gl} & \ding{55} & 7.3 & 11.7 & 84.3 & 147.5 & 26.18 \\
			DMC~\citep{dmc} & \ding{55} & 7.4 & 11.8 & 85.6 & 148.3 & 25.49 \\
			\hline
			P2P~\citep{p2p} & \ding{51} & 6.3 & 9.9 & 85.3 & 154.5 & 2.32 \\
			P2R~(ours) & \ding{51} & \textbf{6.2} & \textbf{9.8} & \textbf{83.3} & \textbf{138.1} & 156.25 \\
			\hline
	\end{tabular}}
	\caption{Comparison of counting losses (100\% Label Pct.)
		\NOTE{are GL and DMC point-based?}\ANS{add a new column to indicate whether it is a point-based method.}
	}
	\label{tab:loss}
\end{table}

\lwn{A theoretical overview about current counting losses is briefly introduced in the first part of Sec.~\ref{sec:reworks}: \textit{Related Works}. Tab.~\ref{tab:loss} presents the empirical comparison, and P2R achieves better performance. Besides, the main paper provides a brief comparison between P2R and P2P in the 2nd paragraph of Sec.~\ref{sec:ssc}, given that both are designed for point-based counting models.}

\begin{figure*}[t]
	\centering
	\includegraphics[width=0.98\textwidth]{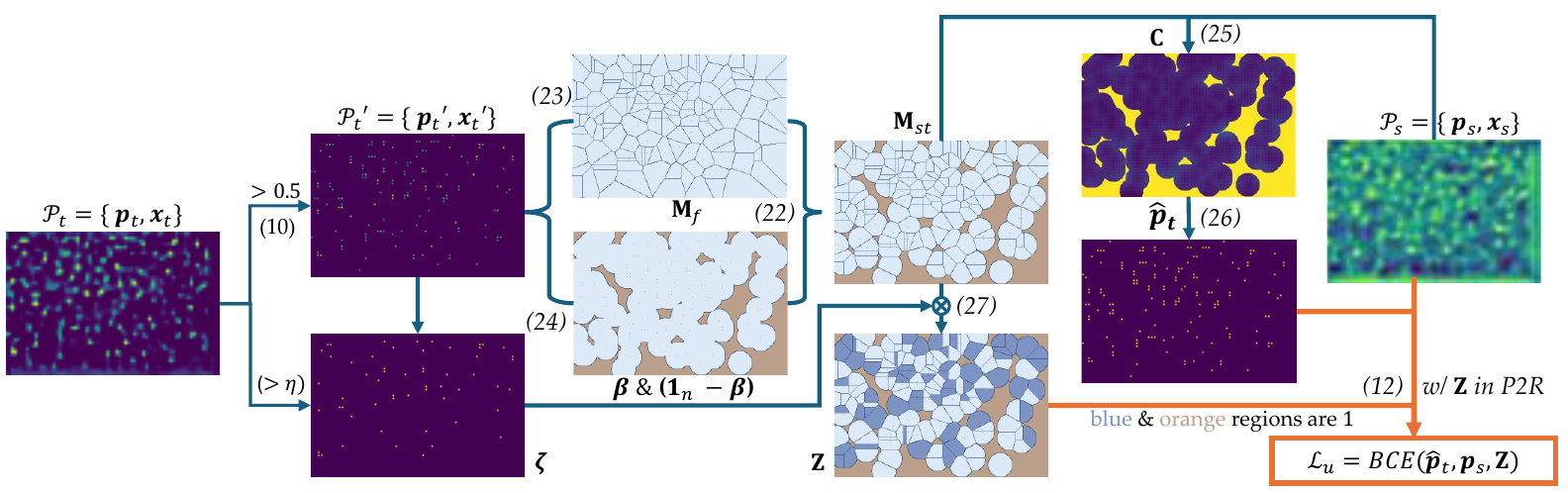}
	\caption{Computation of $\mac{L}_u$ in P2R.}
	\label{fig:pipeline}
\end{figure*}

\begin{figure*}
	\centering
	\includegraphics[width=0.96\textwidth]{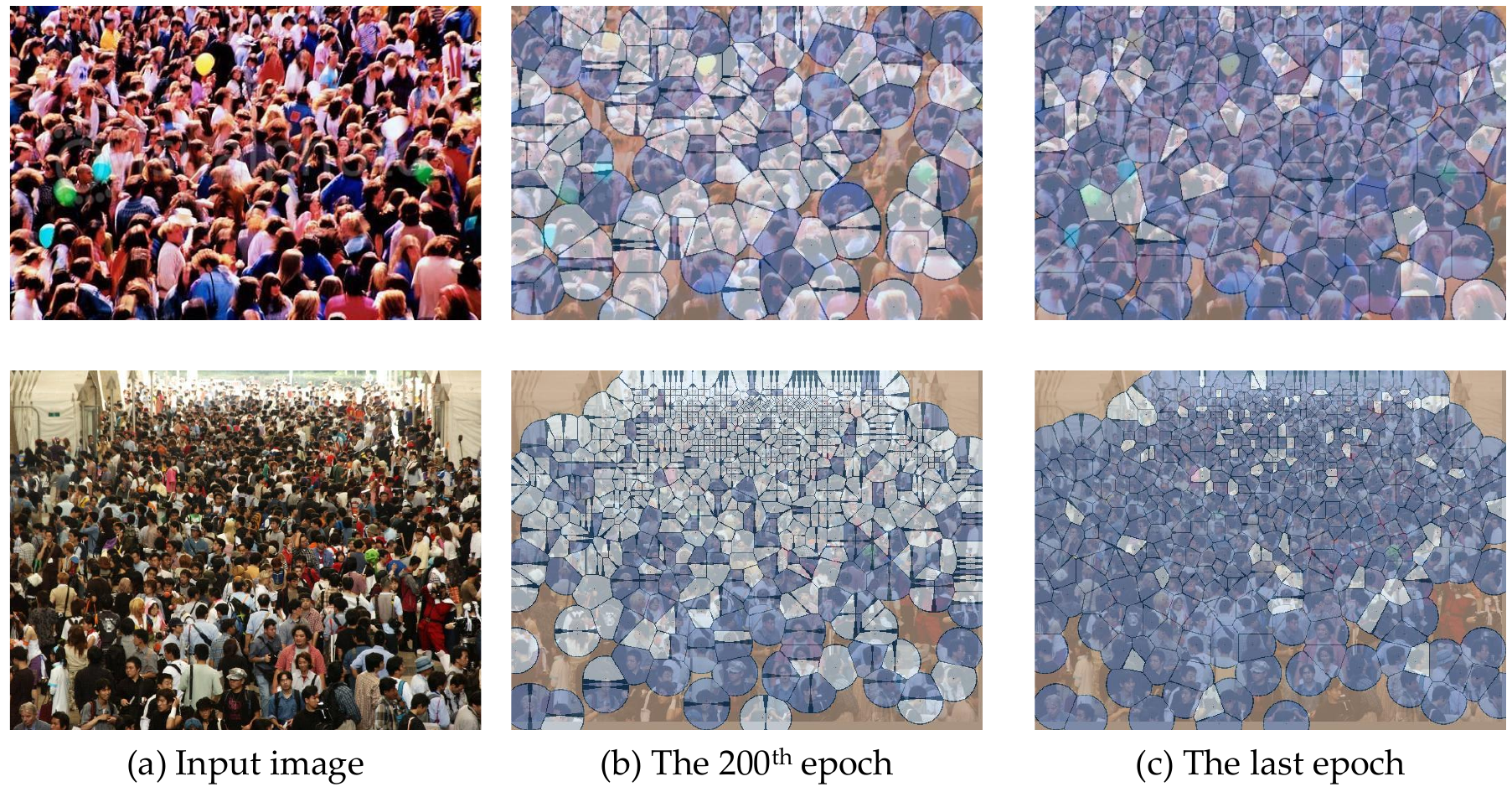}
	\caption{More visualization of the pseudo-labels.}
	\label{fig:loss_vis}
	\vspace{-2mm}
\end{figure*}

\section{Visualization}
\label{sec:vis}

\subsection{Pseudo-Labels}

\lwn{In Fig.~\ref{fig:pipeline}, we illustrate the pipeline of loss computation for P2R. Given the teacher's prediction $\mac{P}_t$, the tensors $\mac{P}_t'$ and $\bsb{\zeta}$ are generated by filtering out pixels with values greater than 0.5 and $\eta$, respectively. Subsequently, the region-to-point matching matrix $\mathbf{M}_{st}$ is constructed via \eqref{eq:new_m}, incorporating the results from the foreground assignment process \eqref{eq:nearestn} and the background definition \eqref{eq:fg}. Consequently, the cost value to determine the learning objective within each region is estimated through \eqref{eq:cost2}. The learning objective $\hat{\bsb{p}} = \hat{\mbf{M}} \mbf{1}$ is then defined, where $\hat{\mbf{M}}$ identifies the potential foreground pixels in each region by \eqref{eq:locmin}. Finally, substituting $\bsb{p}_s$, $\hat{\bsb{p}}_t$, and the trustable region indicator $\mathbf{Z}$ obtained via \eqref{eq:p2r_z} into the BCE loss \eqref{eq:bce_z} yields the final loss value for the P2R loss value in our semi-supervised crowd counting.}

\lwn{In the main paper, we visualize pseudo-labels in Figs.~\ref{fig:p2pr}(e) and \ref{fig:p2pr}(f).  In Fig.~\ref{fig:loss_vis}, we present two more examples to show the evolution of trusted regions from the $200^{\text{th}}$ epoch to the end of training. }

\subsection{Comparison to GT}

\begin{figure*}[t]
	\centering
	\includegraphics[width=0.96\textwidth]{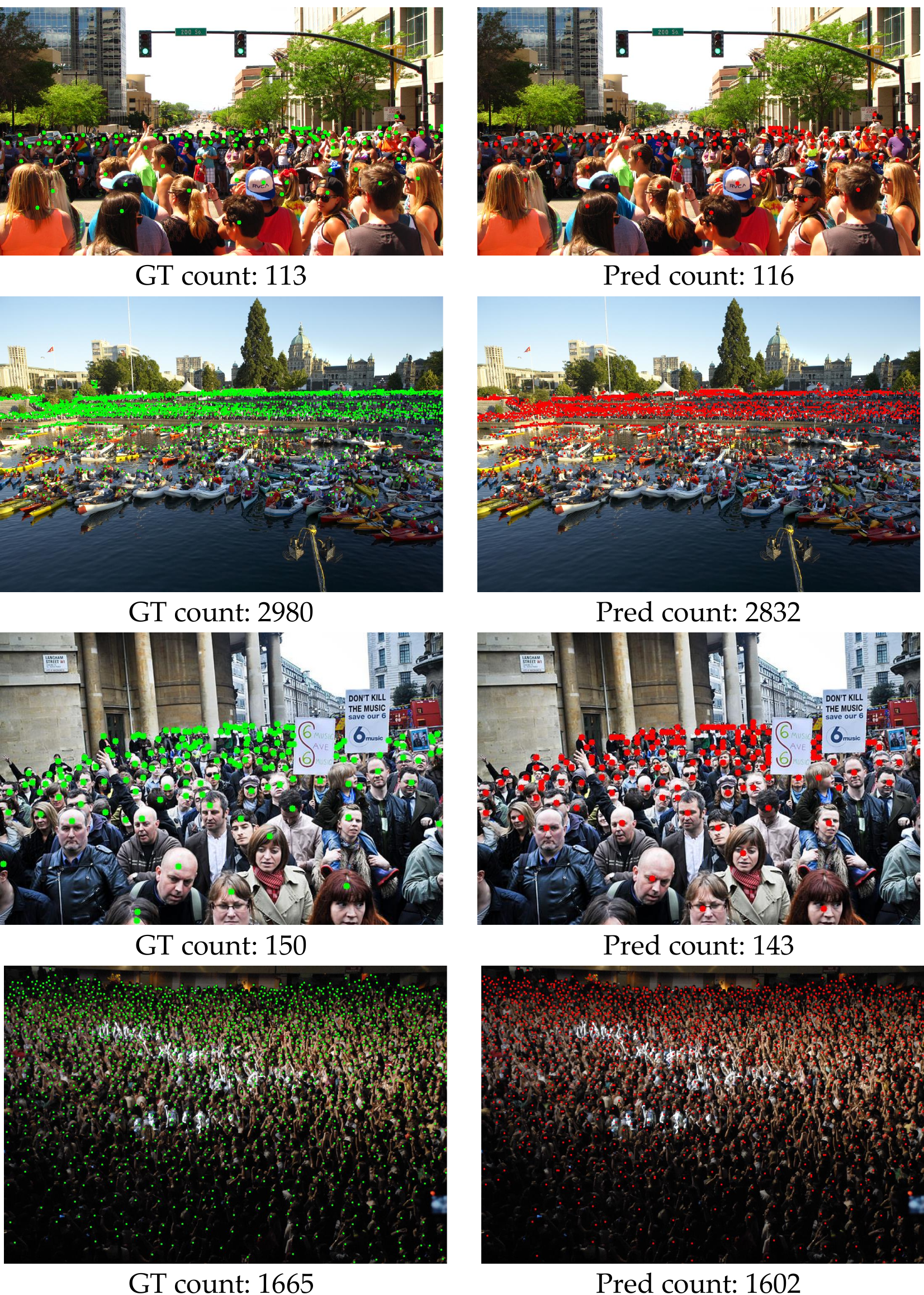}
	\caption{Visualization of P2R's Prediction (\# 1).}
	\label{fig:vis1}
\end{figure*}

In Fig.~\ref{fig:vis1}, 
we present some visualization results of P2R when trained with 40\% labeled data of the UCF-QNRF dataset. P2R can recognize the semantic information and localize pedestrians in the given images effectively.

\begin{figure*}[t]
	\centering
	\includegraphics[width=0.96\textwidth]{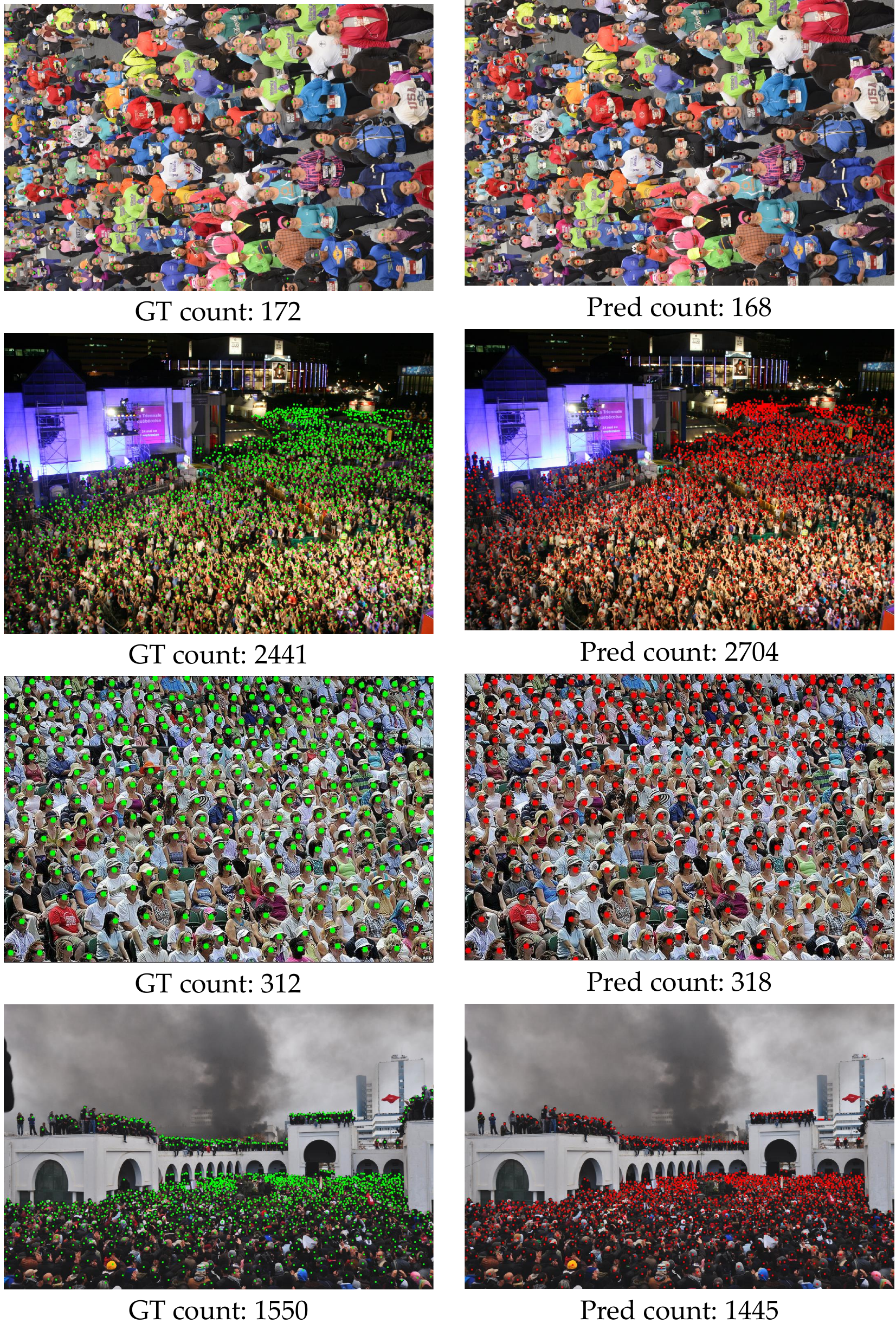}
	\caption{Visualization of P2R's Prediction (\#2)}
	\label{fig:vis2}
\end{figure*}

\end{document}